%% file: colm2026_conference.tex
\documentclass{article} 
\usepackage[preprint]{colm2026_conference}
\usepackage{todonotes}
\usepackage{microtype}
\usepackage{hyperref}
\usepackage{url}
\usepackage{booktabs}
\usepackage{amsmath}
\usepackage{wrapfig}
\usepackage[normalem]{ulem}
\usepackage{booktabs}
\usepackage{multirow}
\usepackage{makecell}
\usepackage{graphicx}

\usepackage{lineno}

\definecolor{darkblue}{rgb}{0, 0, 0.5}
\hypersetup{colorlinks=true, citecolor=darkblue, linkcolor=darkblue, urlcolor=darkblue}

\title{SAW-INT4: System-AWare 4-Bit KV-Cache Quantization for Real-World LLM Serving}

\author{
Jinda Jia\thanks{Equal contribution.}, Jisen Li\footnotemark[1], Zhongzhu Zhou, Jung Hwan Heo, Jue Wang, Tri Dao,\\
\textbf{Shuaiwen Leon Song, Ben Athiwaratkun, Chenfeng Xu, Tianyi Zhang\thanks{Project lead.}, Xiaoxia Wu\footnotemark[2]}
\\
}
\usepackage{underscore}
\begin{document}

\ifcolmsubmission
\linenumbers
\fi

\maketitle

\begin{center}
 \vspace{-7mm}
Together AI \\
\small Code: \url{https://github.com/togethercomputer/saw-int4}
\end{center}

\begin{abstract}

KV-cache memory is a major bottleneck in real-world LLM serving, where systems must simultaneously support latency-sensitive small-batch requests and high-throughput concurrent workloads. Although many KV-cache compression methods improve offline accuracy or compression ratio, they often violate practical serving constraints such as paged memory layouts, regular memory access, and fused attention execution, limiting their effectiveness in deployment.

In this work, we identify the minimal set of 4-bit KV-cache quantization methods that remain viable under these constraints. Our central finding is that a simple design—token-wise INT4 quantization with block-diagonal Hadamard rotation—consistently achieves the best accuracy–efficiency trade-off. Across multiple models and benchmarks, this approach recovers nearly all of the accuracy lost by naive INT4, while more complex methods such as vector quantization and Hessian-aware quantization provide only marginal additional gains once serving compatibility is taken into account.

To make this practical, we implement a fused rotation–quantization kernel that integrates directly into paged KV-cache layouts and introduces zero measurable end-to-end overhead, matching plain INT4 throughput across concurrency levels. Our results show that effective KV-cache compression is fundamentally a systems co-design problem: under real serving constraints, lightweight block-diagonal Hadamard rotation is a viable method that delivers near-lossless accuracy without sacrificing serving efficiency.


\end{abstract}

\input{section1}
\input{section2-small}
\input{section3}
\input{section4}

\bibliography{colm2026_conference}
\bibliographystyle{colm2026_conference}

\appendix
\input{section6}

\input{section5}

\end{document}

%% file: section1.tex
\section{Introduction}

Large language models (LLMs) have made remarkable progress in reasoning~\citep{chain-of-thought},
tool use~\citep{toolllm}, and interaction with complex environments~\citep{react}, driving adoption
across an ever-widening range of applications~\citep{llm-applications}. As these capabilities
advance, long-context inference and agentic workloads have become
increasingly important~\citep{future-agentic-ai}; frontier models now support context windows of
millions or even tens of millions of tokens~\citep{gemini,llama4}. This trend
poses a fundamental challenge for key-value (KV) cache management, whose
memory footprint grows linearly with sequence length~\citep{scaling-transformer}. In production
serving systems, KV cache memory is one of the primary constraints on
throughput~\citep{vllm}, as it directly limits the number of requests that can be
served concurrently on a given set of hardware. Architectural innovations
such as grouped-query attention (GQA)~\citep{gqa} and multi-head latent
attention (MLA)~\citep{mla} reduce the per-token KV footprint relative to
standard multi-head attention (MHA)~\citep{Vaswani+2017}, yet the absolute memory demand
remains substantial at long context lengths. For example, Llama~4 Scout~\citep{llama4}
has a model size of only 218\,GB, but serving a single request at its
maximum context length of 10M tokens requires roughly 1.8\,TiB of KV
cache memory alone---an order of magnitude larger than the model weights
themselves. As context lengths continue to scale and inference-time
compute becomes a more prominent scaling axis~\citep{test-time-compute}, this problem will only
intensify. Consequently, KV cache compression has attracted extensive
research attention~\citep{h2o,kivi,kv-1bit}.

\textbf{The System Gap in Existing KV Cache Compression.}
Despite achieving impressive compression ratios in isolation, existing KV cache compression methods face significant challenges integrating into real LLM serving systems and often fail to deliver end-to-end gains in practice. Production serving engines such as vLLM~\citep{vllm}, SGLang~\citep{sglang}, and TensorRT-LLM~\citep{trtllm} are built around tightly optimized primitives---PagedAttention~\citep{vllm}, continuous batching~\citep{orca}, and FlashAttention kernels~\citep{flashattn}---that leave very little room for additional overhead. Because autoregressive LLM decoding is predominantly memory-bandwidth-bound, even modest extra computation or irregular memory access during the decode phase translates directly into measurable latency increases and throughput degradation. 

Existing compression techniques violate these constraints in several ways. Methods such as KIVI~\citep{kivi} and Kitty~\citep{xia2025kittyaccurateefficient2bit} maintain a fixed-length residual buffer of unquantized key--value pairs alongside quantized tokens, creating a mixed-precision KV cache. However, PagedAttention manages cache memory in fixed-size, uniform-type blocks; accommodating two distinct precisions within the same paged pool requires either fragmented memory layouts or separate page tables, both of which complicate memory management and break the assumptions of existing fused attention kernels. Vector-quantization-based approaches~\citep{commutative-vector-quantization,kv-1bit} face a different but equally problematic bottleneck: encoding and decoding through codebook lookups introduce irregular, data-dependent memory access patterns that are poorly suited to GPU execution and add non-trivial latency to every decoding step. More broadly, techniques that rely on channel-wise quantization, token eviction, or learned codebooks introduce cross-token dependencies or irregular access patterns that are difficult to incorporate into fused FlashAttention kernels and PagedAttention-based memory managers without substantial kernel-level modifications. 

{In summary, many existing methods fundamentally violate the structural and execution constraints of modern LLM serving systems, leading to limited or negative end-to-end benefits despite strong offline accuracy results.} This observation motivates a key question:

\begin{center}
    \textit{Among the many proposed KV-cache compression techniques, which ones actually work under the constraints of real-world serving systems---and how simple can an effective solution be?}
\end{center}

Rather than proposing yet another compression method, this paper takes a step back and conducts a systematic empirical study of serving-compatible INT4 KV-cache quantization. We evaluate a broad spectrum of techniques---na\"ive token-wise INT4, vector quantization with varying codebook sizes, Hessian-aware quantization, orthogonal rotation, and hybrid combinations thereof---all under a unified, system-aware evaluation protocol that jointly measures accuracy and real serving throughput.

Our finding is that Block-Diagonal Rotation (BDR) before token-wise INT4 quantization recovers nearly all accuracy lost by na\"ive quantization. We summarize our contributions:

\begin{itemize}

    \item \textbf{Systematic empirical study of serving-compatible KV-cache quantization.} We identify the key system constraints that practical KV-cache compression must satisfy (Section~\ref{sec:system_constraints}), and conduct a controlled comparison of five families of token-wise quantization techniques---na\"ive INT4, k-means vector quantization, Hessian-aware quantization, Hadamard rotation, and their combinations---across multiple models and benchmarks (Section~\ref{sec:3}).

    \item \textbf{Block-Diagonal Rotation (BDR) is the key enabler; complexity yields diminishing returns.} Our experiments reveal that orthogonal rotation is the single most important ingredient for effective INT4 KV-cache quantization, recovering accuracy to within 1--3 points of BF16 on fragile models where na\"ive INT4 scores zero. More sophisticated methods (k-means with up to 2048 centroids, Hessian-aware calibration) provide only marginal improvements on top of rotation and introduce calibration overhead and deployment complexity that are difficult to justify.

    \item \textbf{Fused implementation with zero serving overhead.} We develop a fused rotation--quantization CUDA kernel based on block-diagonal Hadamard transforms that integrates seamlessly into paged KV-cache layouts and FlashAttention-style decoding. End-to-end serving benchmarks on production workloads confirm that fused INT4 with rotation matches the throughput of plain INT4 within measurement noise, while preserving the latency and throughput advantage of INT4 over BF16.

    \item \textbf{Extensive evaluation across models and tasks.} We validate our findings on Qwen3-4B, Qwen3-8B, Qwen3-32B, and GLM-4.7-FP8 across five reasoning and coding benchmarks (GPQA, HumanEval, LiveCodeBench, AIME25, MATH500), showing that the simplest serving-compatible approach---rotation plus token-wise INT4---consistently provides the best accuracy--efficiency trade-off across model scales.
\end{itemize}
\begin{figure}[t]
    \centering
    \begin{minipage}[t]{0.65\linewidth}
        \centering
        \includegraphics[width=\linewidth]{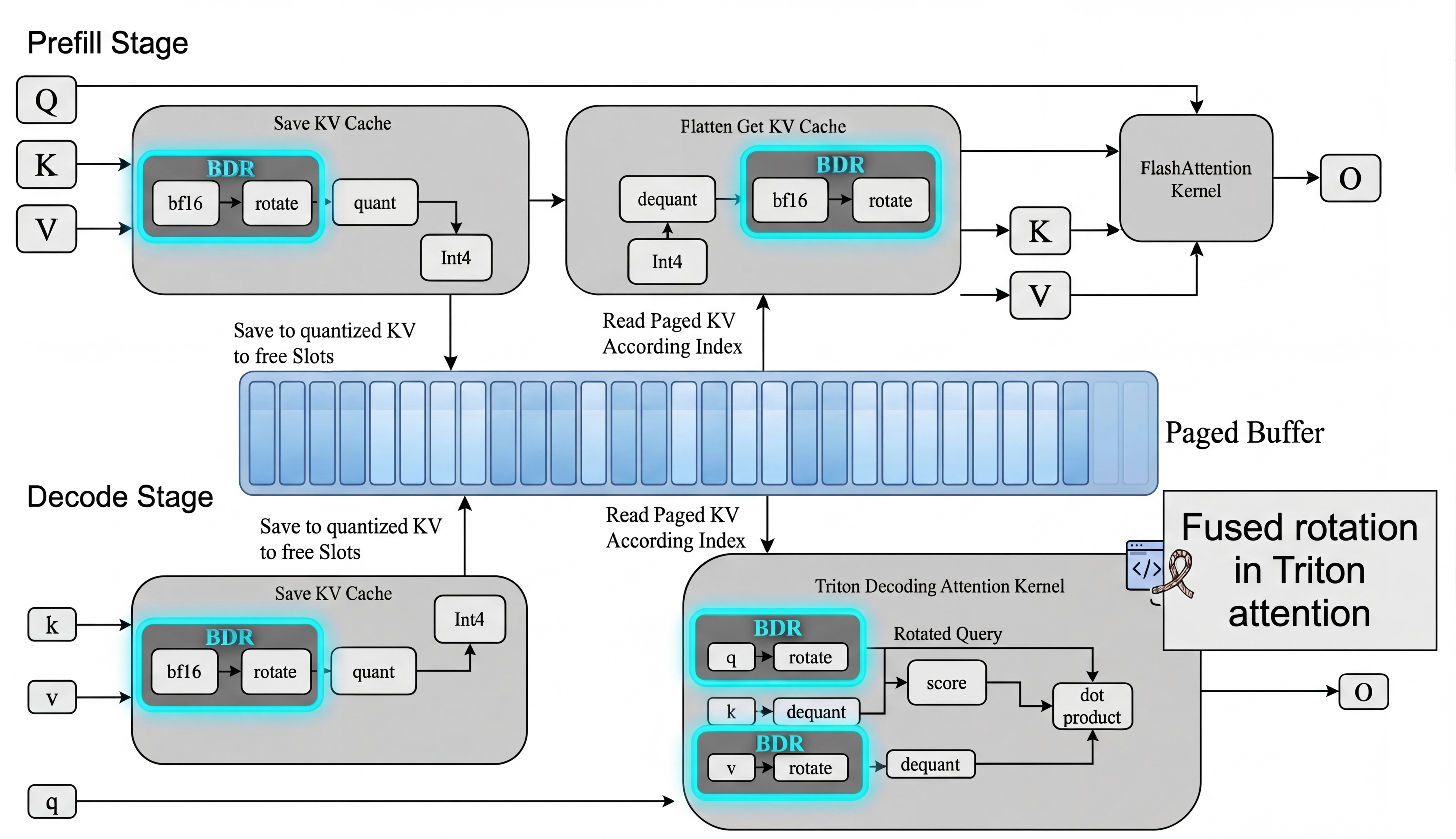}
    \end{minipage}\hfill
    \begin{minipage}[t]{0.34\linewidth}
        \centering
        \includegraphics[width=\linewidth]{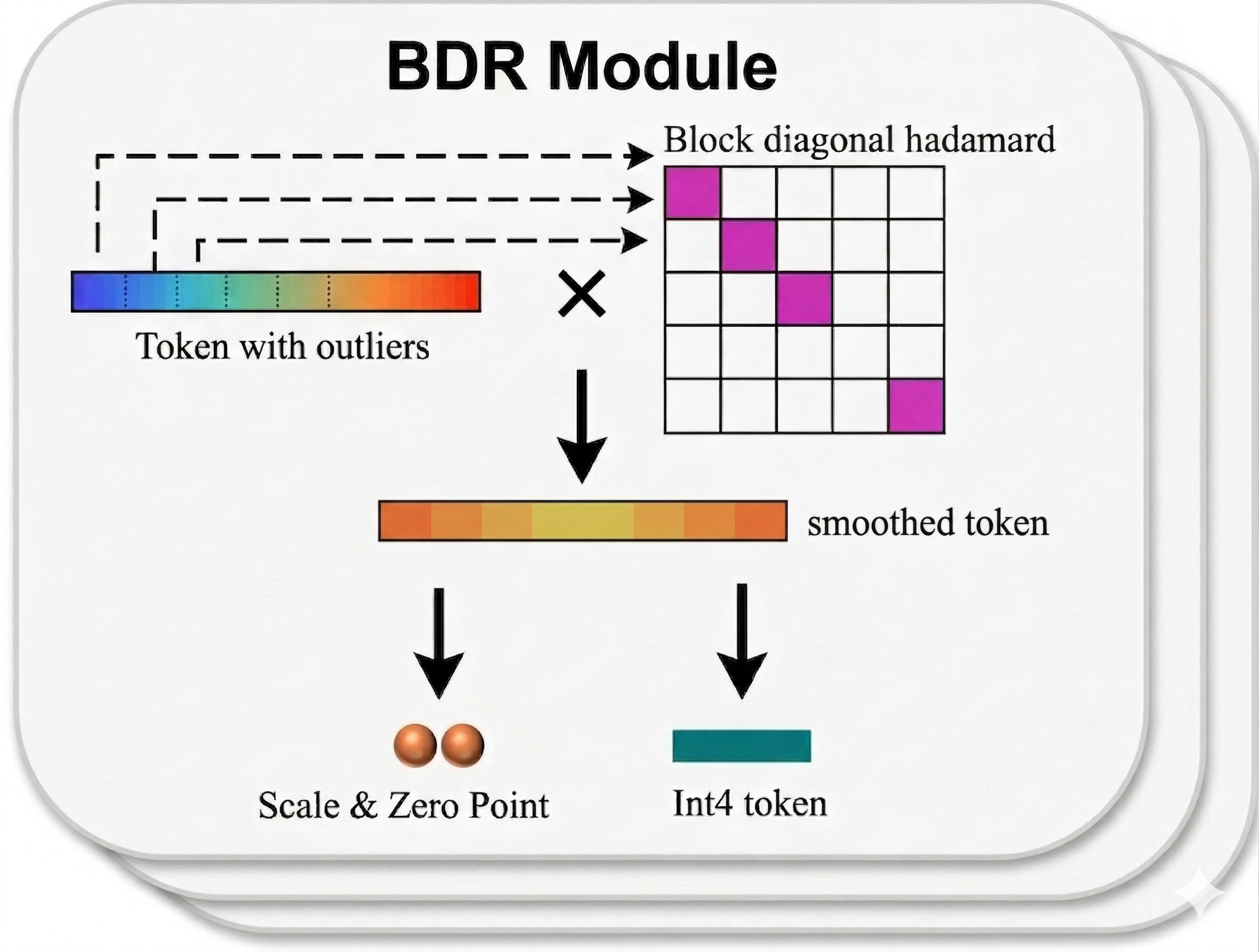}
    \end{minipage}
    \caption{Overview of our system-aware INT4 KV-cache quantization framework (left) and rotate-quantize pipeline (right) for modern LLM serving.}
    \label{fig:overview}
\end{figure}



\begin{table}[t]
\centering
\footnotesize
\caption{Throughput on 2×H100 GPUs with 32 concurrent requests, each with an input length of 8192 tokens, output 1024 tokens. Here, TPS/User denotes tokens/s per user, and TPS/GPU denotes tokens/s per GPU. Acc is the mean computed over 5 tasks; each task is repeated 5 times.}
\label{tab:kv_budget_compare}
\setlength{\tabcolsep}{5pt}
\begin{tabular}{lccccccc}
\toprule
& & \multicolumn{3}{c}{Qwen3-4B-Thinking-2507} & \multicolumn{3}{c}{Qwen3-8B} \\
\cmidrule(lr){3-5} \cmidrule(lr){6-8}
Method & Page Memory & TPS/User & TPS/GPU & Acc. & TPS/User & TPS/GPU  & Acc. \\
\midrule
BF16     & 1$\times$ & 71.7 & 1030.5 & 75.64 &  61.9 &  859.1 & 70.84 \\
INT4     & 4$\times$ & 86.7  & 1217.5 & 0  & 73.4 &  962.8  & 0  \\
\midrule
KMeans C=256  & 4$\times$ & 24.2 & 365.7 & 71.64 & 23.3 & 347.9 & 68.91 \\
Hessian + BDR  & 4$\times$ & 76.8 & 1051.0 & 65.52 & 66.3 & 887.2 & 70.59 \\
\midrule
\textbf{BDR (Ours)} 
          & 4$\times$ & 88.3 & 1242.2 & 73.78 & 73.3 & 986.3 & 69.86 \\
\bottomrule
\end{tabular}
\vspace{1pt}
\begin{minipage}{0.98\linewidth}
\centering
\scriptsize
\begin{flushleft}
\emph{Notes.} BF16 denotes the full-precision KV cache, and INT4 denotes uniform 4-bit quantization. BDR-$k$ applies block-diagonal rotation with block size $k$ before INT4 quantization. K-means refers to vector quantization with $C$ centroids. Hessian+BDR applies block-diagonal rotation followed by Hessian-aware quantization. Naive INT4 fails to produce meaningful outputs. The five evaluation tasks are GPQA-Diamond, HumanEval, LiveCodeBench (v6), AIME25, and MATH500. All results are reported with a 32k sequence length in thinking mode.
\end{flushleft}
\end{minipage}
\end{table}

\subsection{Related Work}

As LLMs support increasingly long context windows, the key-value (KV) cache becomes a primary memory bottleneck. While architectural innovations such as grouped-query attention \citep{gqa} and multi-head latent attention \citep{mla} reduce the per-token footprint, absolute memory demand at long contexts remains a critical challenge, motivating a large body of work on post-training KV cache compression \citep{kv-survey}.

\textbf{KV Cache Compression.} Token eviction methods \citep{h2o, attn-sink} reduce cache size by discarding tokens deemed unimportant based on attention scores or recency. Scalar quantization methods \citep{kivi, hooper2024kvquant, gear, zipcache, xia2025kittyaccurateefficient2bit} reduce the bit-width of KV activations, employing techniques such as per-channel key quantization, mixed-precision schemes, and outlier-aware encoding to achieve 2–4 bit compression with minimal quality loss. Beyond scalar approaches, vector quantization methods \citep{kv-1bit, li2025commvq, turboquant} encode multiple dimensions of KV vectors jointly, exploiting inter-channel dependencies to push compression to 1–3 bits per channel while preserving model quality. Orthogonally, low-rank decomposition methods \citep{low-rank-kv, kv-low-rank-proj} project keys and values into lower-rank subspaces, reducing both cache size and attention latency.



\textbf{Quantization Techniques.} KV cache quantization methods draw on insights from broader LLM quantization research. A central challenge is activation outliers: LLM.int8()~\citep{LLM-int8} isolates high-magnitude outlier dimensions into higher precision, while SmoothQuant~\citep{smoothquant} migrates quantization difficulty from activations to weights via per-channel scaling. For weight-only compression, Hessian-aware methods such as GPTQ~\citep{frantar2022gptq} and Happi~\citep{kim2026happi} use second-order information to guide rounding and mixed-precision assignment. A separate line of work removes outliers through rotation: QuaRot~\citep{ashkboos2024quarot} applies randomized Hadamard transforms to produce near-uniform distributions, and SpinQuant~\citep{spinquant} learns optimal rotations on a Stiefel manifold, closing much of the accuracy gap at 4-bit quantization of weights, activations, and KV caches. At the extreme end, vector quantization methods such as QuIP\#~\citep{quip-sharp} and AQLM~\citep{aqlm} jointly encode groups of weight values using structured or learned codebooks, achieving strong results at 2--3 bits.

%% file: section2-small.tex
\section{LLM Serving System and KV Compression Constraints}
\label{sec:system_constraints}

\vspace{-0.2cm}
A practical KV-cache quantization method must be evaluated not only by its offline compression ratio, but by whether it preserves throughput in a real serving stack. LLM serving imposes two non-negotiable constraints that invalidate  some compression paradigms.

\textbf{Paged Memory Layouts:} Modern systems \citep[e.g.,][]{vllm, sglang} divide the KV cache into fixed-size, non-contiguous blocks to enable dynamic batching. This architectural reality breaks several approaches:
\begin{itemize}
    \item \textit{Token Eviction:} Evicting ``unimportant'' tokens \citep[e.g.,][]{h2o} does not reduce the physical memory footprint unless an entire block is cleared, causing massive internal fragmentation.
    \item \textit{Mixed-Precision:} Methods using varying bit-widths \citep[e.g.,][]{xia2025kittyaccurateefficient2bit} break the uniform layout required for efficient page table indexing.
    \item \textit{Channel-Wise Scaling:} Because tokens span non-contiguous blocks, computing shared scaling factors across channels \citep[e.g.,][]{kivi} requires highly inefficient cross-block memory access.
\end{itemize}

\begin{wrapfigure}{r}{0.5\textwidth} 
   \vspace{-0.94cm}
    \centering
    \includegraphics[width=0.77\linewidth]{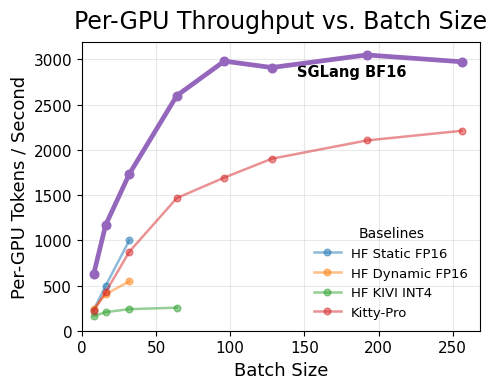}
    \vspace{-0.2cm}
    \caption{\footnotesize Qwen3-8B Per-GPU throughput vs. batch size. More complex methods (e.g., Kitty) do not match the performance of unquantized SGLang BF16. All non-SGLang-BF16 throughput numbers are measured with Hugging Face \textit{model.generate}. which lacks continuous batching and PagedAttention, underestimating their performance.} 
    \label{fig:throughput_gap}
    \vspace{-0.6cm} 
\end{wrapfigure}

While methods like Kitty achieve strong offline compression, their complex memory access patterns make integration into optimized paged-memory engines highly non-trivial. As shown in Figure \ref{fig:throughput_gap}, when deployed in standard environments (e.g., Hugging Face generate) that lack PagedAttention and continuous batching, their end-to-end throughput falls massively behind even unquantized baselines running on highly optimized serving engines like SGLang. This highlights that quantization cannot be evaluated in a vacuum; if a compression method's complexity precludes the use of standard system optimizations, it is a net negative for deployment.

\textbf{Attention Kernel:} High-throughput decoding relies on FlashAttention style kernels that read directly from the paged buffer; therefore, dequantization \textit{must} be fused in-kernel~\citep{flashattn}. For a decoding step:$o = \mathrm{softmax}(qK^\top / \sqrt{d})V
$ where $q$ is a single query token and $(K, V)$ are read from the paged KV cache. This strict execution precludes:
\begin{itemize}
    \item \textit{Codebook Lookups:} Vector quantization \citep[e.g.,][]{commutative-vector-quantization} requires global memory lookups during dequantization, increasing latency and bandwidth pressure.
    \item \textit{Basis-Transforms:} PCA/SVD methods require on-the-fly matrix-vector multiplications, inflating register pressure and destroying kernel tiling efficiency.
    \item \textit{Strided Memory Access:} Channel-wise quantization often forces channel-major memory access, destroying the memory coalescing required by GPU attention kernels.
\end{itemize}

\textbf{The Token-Wise Imperative:} Taken together, these system constraints leave one most viable path. Token-wise quantization naturally operates independently on each token, avoids cross-block dependencies, and preserves coalesced memory access patterns. It is the only paradigm inherently compatible with paged layouts and fused kernels. Having established token-wise quantization as the necessary foundation, the remaining challenge is recovering the accuracy lost by naive INT4 implementations that we deconstruct in Section \ref{sec:3}.

%% file: section3.tex
\section{Method Comparison for Rotation Design Intuition}
\label{sec:3}

To isolate the impact of different design choices under practical LLM serving constraints (e.g., paged KV-cache and fused kernels), we conduct a controlled token-wise comparison of four quantization strategies: naive INT4, vector quantization (KMeans)~\citep{flashkmeans2026,xi2026quantvideogen}, Hessian-aware quantization~\citep{kim2026happi,zhou2026care}, and Hadamard rotation-based quantization. This section highlights our core design intuition; detailed derivations and extended evaluations are deferred to Appendix~\ref{sec:appendix-methods}.


\textbf{Token-Wise INT4 Quantization.} We adopt an asymmetric token-wise INT4 scheme, quantizing each head vector $x_{t,h}$ using a per-token and per-head scale and zero-point. While highly system-friendly---reducing the KV-cache footprint from BF16 by $4\times$ while preserving tensor layout---naive INT4 suffers from severe accuracy degradation. This is primarily driven by \textbf{channel-wise range heterogeneity}: high-magnitude outliers in specific channels artificially expand the quantization grid, leading to massive quantization errors across the rest of the vector~\citep{hooper2024kvquant,kivi,kv-1bit}.

\textbf{Block-Diagonal Rotation Quantization.}
To remove data-driven calibration while mitigating channel-wise outliers, we apply an orthonormal transform $H$ to rotate the KV-cache~\citep{ashkboos2024quarot,tseng2024quipsharp}:
\begin{equation}
\tilde K = \mathrm{quant}(K H), \quad \tilde V = \mathrm{quant}(V H).
\end{equation}
As $H$ is orthogonal, it preserves the $\ell_2$ norm while redistributing energy across dimensions, reducing channel-wise variance and smoothing outliers—the main source of degradation under low-bit quantization (e.g., INT4). Attention can be computed directly in the rotated space or equivalently via $H^\top$, incurring no additional approximation error.

In practice, we use randomized Hadamard transforms. However, full dense rotation over the full hidden dimension is costly. We therefore introduce \emph{block-diagonal rotation (BDR)}, decomposing the global transform into independent local rotations. Let $d_h$ be the head dimension. We partition each vector into blocks of size $h$:
\begin{equation}
K = [K^{(1)}, \dots, K^{(d_h/h)}], \quad
V = [V^{(1)}, \dots, V^{(d_h/h)}], \quad K^{(i)}, V^{(i)} \in \mathbb{R}^h.
\end{equation}
Each block is independently rotated and quantized:
$\tilde K^{(i)} = \mathrm{quant}(K^{(i)} H_h),
\tilde V^{(i)} = \mathrm{quant}(V^{(i)} H_h) $
where $H_h \in \mathbb{R}^{h \times h}$ is a Hadamard matrix. This induces a block-diagonal transform $H_{\text{blk}} = \mathrm{diag}(H_h, \dots, H_h)$, approximating a full rotation at significantly lower cost and with better kernel efficiency. The block size $h$ controls the trade-off: larger $h$ improves cross-channel mixing and robustness, while smaller $h$ favors parallelism and kernel fusion.

While smaller $h$ reduces per-rotation compute and improves kernel-level efficiency, 
we find that these gains do not consistently translate to end-to-end throughput improvements. 
This is because autoregressive decoding is primarily memory-bound, and rotation is not the dominant bottleneck. 
As a result, overly small blocks sacrifice quantization quality without delivering meaningful system-level gains. We will explain this in Section \ref{sec:4}.

\textbf{Hessian-Aware Quantization.} Following and inspired by prior Hessian- (or covariance-) aware quantization approaches~\citep{kim2026happi,zhou2026care}, we use decode-time query statistics to estimate attention-sensitive directions and derive a learned per-layer rotation for KV quantization. We instrument the serving engine to collect query vectors during offline calibration and, for each layer $\ell$, form the second-moment matrix $M_\ell = \frac{1}{N_\ell}\sum_{i=1}^{N_\ell} q_i q_i^\top$, which yields the score-error objective $\mathcal{L}_\ell = \delta k^\top M_\ell \delta k$ for key quantization error $\delta k$. We then take the resulting orthogonal eigenvector matrix $R_\ell$ as the learned per-layer rotation. At inference time, $R_\ell$ is composed with a Hadamard transform and applied to both $Q$ and $K$, preserving full-precision attention logits while shifting quantization noise away from sensitive directions. 

\textbf{Vector Quantization (KMeans).} Vector quantization mitigates outlier degradation by mapping head vectors to a learned codebook of centroids. Here, $C$ denotes the number of centroids (codebook size) e.g., ``KM w $C{=}256$'' means KMeans with 256 clusters. By clustering similar activations rather than uniformly slicing the vector space, VQ recovers a substantial portion of the accuracy lost by scalar INT4. However, performance gains are not strictly monotonic with codebook size, highlighting a trade-off between representational capacity and the stability of the clustering algorithm during offline calibration. 
Detailed algorithmic description and residual-quantization formulation are provided in Appendix~\ref{sec:appendix-methods}.

\begin{table}[h!]
\centering
\footnotesize
\setlength{\tabcolsep}{4pt}
\caption{Expanded comparison on Qwen3-4B-thinking-2507. Entries are $\mu \pm \sigma$ over 5 runs. Mean is computed over available task means (all 5 tasks where available).}
\label{tab:kv_compare}
\begin{tabular}{l c c c c c c c}
\toprule
\textbf{Method} & \textbf{GPQA} & \textbf{HumanE} & \textbf{LCB v6} & \textbf{AIME} & \textbf{MATH} & \textbf{Mean} & \textbf{Drop} \\
\midrule
BF16 & 67.27$\pm$1.80 & 94.05$\pm$0.54 & 48.66$\pm$2.20 & 74.67$\pm$1.83 & 93.55$\pm$0.33 & 75.64 & 0.00 \\
INT4 & 0.00$\pm$0.00 & 0.00$\pm$0.00 & 0.00$\pm$0.00 & 0.00$\pm$0.00 & 0.00$\pm$0.00 & 0.00 & -75.64 \\
\midrule
BDR-16    & 50.61$\pm$2.35 & 56.20$\pm$2.40 & 26.20$\pm$2.85 & 54.67$\pm$4.47 & 86.45$\pm$0.61 & 54.83 & -20.81 \\
BDR-64     & 63.13$\pm$1.38 & 89.10$\pm$0.39 & 46.66$\pm$1.33 & 69.34$\pm$4.35 & 93.23$\pm$0.52 & 72.29 & -3.35 \\
BDR-128  & 66.37$\pm$2.19 & 89.78$\pm$0.80 & 46.20$\pm$1.94 & 70.00$\pm$4.08 & 93.19$\pm$0.51 & 73.11 & -2.53 \\
BDR-128 (K)       & 65.25$\pm$2.12 & 90.49$\pm$0.45 & 48.54$\pm$1.65 & 71.33$\pm$5.06 & 93.27$\pm$0.36 & 73.78 & -1.86 \\
\midrule
Hessian+BDR-128 & 60.10$\pm$2.03 & 84.27$\pm$1.15 & 42.11$\pm$1.96 & 53.33$\pm$4.85 & 87.80$\pm$0.42 & 65.52 & -10.41 \\
\midrule
KM w C=1  & 54.14$\pm$1.53 & 81.76$\pm$0.97 & 31.58$\pm$3.15 & 18.67$\pm$3.80 & 81.92$\pm$0.70 & 53.61 & -22.03 \\
KM w C=16 & 61.52$\pm$2.43 & 91.00$\pm$0.75 & 42.81$\pm$2.53 & 64.00$\pm$3.65 & 93.07$\pm$0.30 & 70.48  & -5.16 \\
KM w C=256  & 63.94$\pm$0.58 & 91.85$\pm$0.51 & 41.29$\pm$1.73 & 68.00$\pm$5.05 & 93.11$\pm$0.11 & 71.64 & -4.00 \\
KM w C=2048 & 60.71$\pm$5.09 & 93.17$\pm$0.49 & 43.16$\pm$4.59 & 74.00$\pm$6.41 & 93.03$\pm$0.43 & 72.81 & -2.83 \\
KM + BDR-128  & 65.25$\pm$0.49  & 93.27$\pm$0.63 & 47.25$\pm$1.63 & 67.33$\pm$2.5 & 93.63$\pm$0.51 & 73.35 & -2.29 \\
\bottomrule
\end{tabular}

\begin{flushleft}
{\scriptsize
\emph{Notes.}
BDR-$k$ refers to Block-Diagonal Rotation with block size $k$, applied prior to INT4 quantization on Key and Value. BDR-$k$ (K) means we only apply rotation on K but not V before their quantization.
Hessian+BDR-128 denotes applying BDR-128 followed by Hessian-based quantization.
KM ($C$) denotes K-means (vector quantization) with $C$ centroids.
KM + BDR-128 applies BDR-128 on top of KM with $C=2048$ (the KM-only result for $C=2048$ is shown above; other variants are omitted due to space constraints).
}
\end{flushleft}
\vspace{-0.6cm}
\end{table}

\begin{table*}[h]
\vspace{-0.2cm}
\centering
\footnotesize
\caption{Expanded comparison on Qwen3-8B and GLM-4.7 (358B).  Entries are $\mu \pm \sigma$.}
\label{tab:kv_compare_more}
\begin{tabular}{l c c c c c c c}
\toprule
\textbf{Method} & \textbf{GPQA} & \textbf{HumanE} & \textbf{LCB v6} & \textbf{AIME} & \textbf{MATH} & \textbf{Mean} & \textbf{Drop} \\
\midrule
\multicolumn{8}{c}{\textbf{Qwen3-8B}} \\
BF16                   & 56.67$\pm$2.30 & 85.95$\pm$1.01 & 49.01$\pm$2.13 & 70.00$\pm$3.33 & 92.59$\pm$0.62 & 70.84 & -- \\
INT4            & 0.00$\pm$0.00 & 0.00$\pm$0.00 & 0.00$\pm$0.00 & 0.00$\pm$0.00 & 0.00$\pm$0.00 & 0.00 & -- \\
BDR-16    & 54.85$\pm$2.44 & 86.05$\pm$0.80 & 47.13$\pm$1.52 & 59.33$\pm$5.96 & 92.06$\pm$0.86 & 67.88 & -2.96 \\
BDR-64     & 54.55$\pm$1.55 & 87.59$\pm$0.48 & 47.48$\pm$2.16 & 64.00$\pm$4.35 & 92.18$\pm$0.64 & 69.16 & -1.68 \\
BDR-64 (K)        & 56.97$\pm$0.83 & 86.59$\pm$0.98 & 47.02$\pm$1.58 & 66.00$\pm$4.94 & 92.71$\pm$0.37 & 69.86 & -0.99 \\
BDR-128    & 54.85$\pm$3.17 & 86.44$\pm$0.42 & 47.95$\pm$2.15 & 68.00$\pm$2.98 & 92.63$\pm$0.27 & 69.97 & -0.87 \\

\midrule
\multicolumn{8}{c}{\textbf{GLM-4.7 (358B)}} \\

BF16                   & 73.23$\pm$1.33 & 91.46$\pm$0.65 & 49.12$\pm$0.59 & 80.00$\pm$3.33 & 95.66$\pm$0.61 & 77.89 & -- \\
INT4             & 73.23$\pm$1.34 & 91.14$\pm$0.26 & 45.81$\pm$1.79 & 80.00$\pm$0.00 & 95.86$\pm$0.31 & 77.21 & -0.68 \\
BDR-16    & 72.90$\pm$0.29 & 91.30$\pm$0.63 & 50.68$\pm$0.89 & 78.89$\pm$1.92 & 95.46$\pm$0.42 & 77.85 & -0.04 \\
BDR-64    & 70.87$\pm$1.54 & 92.11$\pm$0.31 & 49.90$\pm$1.79 & 81.11$\pm$3.85 & 95.59$\pm$0.35 & 77.92 & +0.03 \\
BDR-128    & 73.74$\pm$1.01 & 91.30$\pm$0.92 & 50.49$\pm$1.47 & 78.89$\pm$1.92 & 95.32$\pm$0.12 & 77.95 & +0.06 \\

\bottomrule
\end{tabular}
\vspace{-0.5cm}
\end{table*}

\subsection{Empirical Results for Design Intuition}
\label{subsec:empirical}
We compare prior token-wise compatible KV-cache compression methods under a unified protocol on \texttt{Qwen3-4B-thinking-2507}~\citep{Qwen3}. Results are reported as $\mu \pm \sigma$ over 5 runs on GPQA~\citep{rein2024gpqa}, HumanEval~\citep{chen2021evaluating}, LiveCodeBench v6~\citep{jain2024livecodebench}, AIME25~\citep{AIME25}, and MATH500~\citep{hendrycks2021measuring}. We use $T=0.6$, max sequence length 32k, and consistent decoding settings across methods.

\textbf{Rotation recovers most of the INT4 degradation.} As shown in Table~\ref{tab:kv_compare}, naïve INT4 quantization is not viable on Qwen3-4B: all reported metrics collapse to nearly zero, indicating that low-bit KV-cache compression alone fails under realistic decoding conditions. Block-diagonal rotation, however, restores most of the lost accuracy. Increasing the block size from 16 to 64 yields a substantial improvement in mean score (54.83 $\rightarrow$ 72.29), whereas increasing it further to 128 brings only a modest additional gain (73.11 versus 75.64 for BF16). These results suggest that moderate block-wise mixing is already sufficient to suppress the dominant channel-wise outliers induced by quantization.

We further observe, consistent with Kitty~\citep{xia2025kittyaccurateefficient2bit}, that applying rotation only to the key cache is sufficient in practice. Specifically, BDR-128 (K) in Table~\ref{tab:kv_compare} and BDR-64 (K) in Table~\ref{tab:kv_compare_more} achieve accuracy comparable to rotating both K and V. Section~\ref{sec:4} complements this accuracy analysis with a runtime study, showing that a block size of 128  offers the best overall trade-off between quality and efficiency.


\textbf{Effect of quantization complexity.} More complex quantization schemes do not substantially improve on this behavior. Vector quantization with KMeans benefits from larger codebooks, improving from 53.61 at $C{=}1$ to 72.81 at $C{=}2048$, but still remains slightly below BDR-128. Hessian-aware quantization performs markedly worse than BDR-128 (63.23 versus 73.11), despite requiring extra calibration and offline complexity. Combining KMeans with rotation gives the strongest compressed result on Qwen3-4B (73.35), but the gain over BDR-128 alone is small. Overall, once rotation is applied, the remaining headroom is limited, and increasing quantization complexity appears to yield diminishing returns. For more combination results, see Table~\ref{tab:kv_compare1}  and Table~\ref{tab:centroid_ablation} in  Appendix~\ref{sec:ablation}.

\textbf{Scaling behavior across model sizes.} Table~\ref{tab:kv_compare_more} shows the same overall trend on \texttt{Qwen3-8B}: naïve INT4 again collapses, whereas rotation-based INT4 remains nearly lossless. In particular, BDR-128 achieves 69.97, only 0.87 points below the BF16 baseline of 70.84. \texttt{GLM-4.7} exhibits a different pattern. Even naïve INT4 stays close to BF16 (77.21 vs.\ 77.89), suggesting that this model is intrinsically more robust to low-bit KV-cache quantization. Rotation still provides a further, albeit smaller, improvement.

Taken together, these results suggest that the need for sophisticated compression is strongly model-dependent, but that block-diagonal rotation is the most reliable and system-compatible mechanism across settings. For sensitive models such as Qwen3, it turns an otherwise unusable INT4 baseline into a near-lossless one; for more robust models such as GLM-4.7, it preserves accuracy essentially for free.

%% file: section4.tex
\vspace{-0.2cm}
\section{Serving-Compatible KV Compression Method}
\vspace{-0.2cm}
\label{sec:4}

Having established that block-diagonal Hadamard rotation (BDR) recovers most of the accuracy lost by token-wise INT4 quantization, we now evaluate whether it preserves the serving efficiency critical to production LLM systems. A key observation is that kernel microbenchmarks and end-to-end serving performance need not align: plain INT4 can achieve higher isolated kernel bandwidth by avoiding rotation entirely, yet fused rotation introduces little measurable end-to-end penalty. This section explains why.

\subsection{Kernel Optimization in Implementation}
\label{sec:kernel_opt}

Figure~\ref{fig:overview} illustrates how we integrate rotation into decoding. In a Triton decode kernel, each program processes one query and streams tiled key/value blocks for a single decoding step. Let $q$ denote the current query, and let $\tilde{K}_{\mathrm{rot}} \in \mathbb{R}^{S \times h}$ denote the INT4 key cache stored in the rotated space, where $S$ is the context length and $h$ is the rotation block size. The pre-softmax logits are computed as
$
(qH)\,\mathrm{dequant}(\tilde{K}_{\mathrm{rot}})^\top,
$
 with the Hadamard matrix $H$.

A key systems observation is that the cost of rotation is dominated less by arithmetic than by data movement. Applying rotation in a separate pass over the key cache would introduce extra kernel launches and additional global-memory traffic over a tensor of size $S \times R$, which is particularly undesirable during decoding since attention is already memory-bandwidth bound. To avoid this overhead, we fuse rotation directly into the decode attention kernel. Because the kernel already streams query, key, and value tiles, the query can be rotated on the fly and consumed immediately by the $qK$ dot-product loop, without materializing intermediate results. This design preserves the efficiency of the original INT4 decode path while avoiding an extra pass over KV-cache data.



Table~\ref{tab:decode_kernel_profile} reports the decode-time profiling breakdown. INT4-Fused-RotateK closely matches plain INT4 in total kernel time, showing that most rotation overhead is hidden once fused into the decode path. By contrast, the unfused implementation incurs additional overhead from separate rotation-related work. This confirms that the main systems objective is to eliminate extra passes over KV-cache data rather than to minimize the arithmetic cost of rotation in isolation. Detailed prefilling and decoding breakdowns are in Appendix~\ref{sec:appendix-kernel-profiling}.


\begin{table}[h]
\centering
\vspace{-.2cm}
\small
\setlength{\tabcolsep}{2pt}
\caption{Decode kernel profiling breakdown by category for Qwen3-32B on 2$\times$H100 with $\mathrm{tp}=2$ at batch size 32 (single decoding step). Fused rotation closely matches plain INT4, while unfused rotation introduces additional overhead from extra auxiliary work.}
\label{tab:decode_kernel_profile}
\begin{tabular}{l|cc|cc|cc}
\toprule
\textbf{Category} & \multicolumn{2}{c|}{\textbf{INT4}} & \multicolumn{2}{c|}{\textbf{INT4-Fused-RotateK}} & \multicolumn{2}{c}{\textbf{INT4-Unfused-RotateK}} \\
 & \textbf{Time (ns)} & \textbf{Pct (\%)} & \textbf{Time (ns)} & \textbf{Pct (\%)} & \textbf{Time (ns)} & \textbf{Pct (\%)} \\
\midrule
MatMul & 176,800 & 33.16 & 176,703 & 33.35 & 175,391 & 32.44 \\
Attention & 326,842 & 61.30 & 322,880 & 60.93 & 322,816 & 59.71 \\
Comm & 11,200 & 2.10 & 11,519 & 2.17 & 14,496 & 2.68 \\
\texttt{\_quantized\_set\_kv\_int4\_kernel} & 5,504 & 1.03 & 6,112 & 1.15 & 5,504 & 1.02 \\
Norm\_Act\_RoPE & 12,800 & 2.40 & 12,704 & 2.40 & 13,026 & 2.41 \\
Outside\_Kernel\_Rotation & -- & -- & -- & -- & 9,376 & 1.73 \\
\midrule
\textbf{TOTAL} & \textbf{533,146} & \textbf{100.00} & \textbf{529,918} & \textbf{100.00} & \textbf{540,609} & \textbf{100.00} \\
\bottomrule
\end{tabular}
\begin{flushleft}
{\scriptsize
\textit{Notes.}
MatMul: GEMM kernels; 
Attention: decode-attention kernels; 
Comm: \texttt{cross\_device\_reduce\_*}; 
\texttt{\_quantized\_set\_kv\_int4\_kernel}: INT4 KV-cache write (including fused Hadamard + KV write); 
Norm\_Act\_RoPE: RMSNorm, activation, RoPE, and QK-norm; 
Outside\_Kernel\_Rotation: unfused rotation overhead.
}
\end{flushleft}
   \vspace{-0.5cm}
\end{table}

\subsection{From Kernel Throughput to End-to-End Latency}
\label{sec:e2e_analysis}
\begin{wrapfigure}{r}{0.5\textwidth}
   \vspace{-0.5cm}
    \centering
  \includegraphics[width=1.\linewidth]{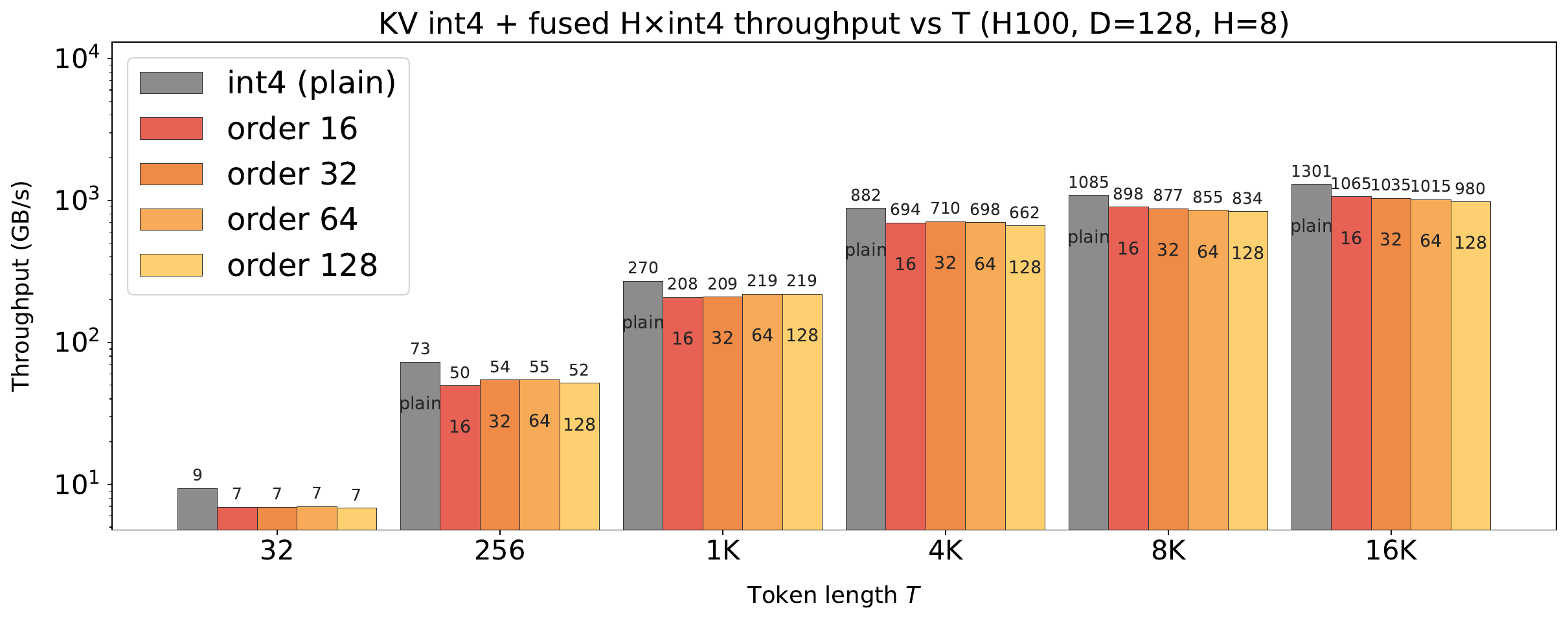}
  \caption{\footnotesize Effective bandwidth (GB/s) on H100 for the fused block rotate--quantize--save KV-cache kernel, grouped by sequence length, with $Head Dimension{=}128$ and $Heads{=}8$. Within each group, bars correspond to different Hadamard block orders $h\in\{16, 32, 64, 128\}$.}
  \label{fig:kv_kernel_gbps}
    \vspace{-0.5cm}
\end{wrapfigure}
As expected, Figure~\ref{fig:kv_kernel_gbps} shows that plain INT4 achieves the highest isolated kernel bandwidth since it avoids rotation entirely. Consequently, fused rotation is slightly slower in this microbenchmark. However, this gap should not be overinterpreted: it reflects only the local KV-cache kernel cost, not the  serving stack.

In end-to-end serving, latency depends not only on the decode kernel, but also on scheduling, memory management, communication, and other runtime overheads  (as displayed in Table~\ref{tab:decode_kernel_profile}). Moreover, the rotated INT4 path accounts for only a fraction of total decode latency, and the incremental cost of rotation is smaller still once integrated into the normal query-key streaming loop. As a result, the kernel-level advantage of INT4 largely disappears in end-to-end latency.

\begin{figure*}[t]
    \centering
    \begin{minipage}[t]{0.48\textwidth}
        \centering
        \includegraphics[width=\linewidth]{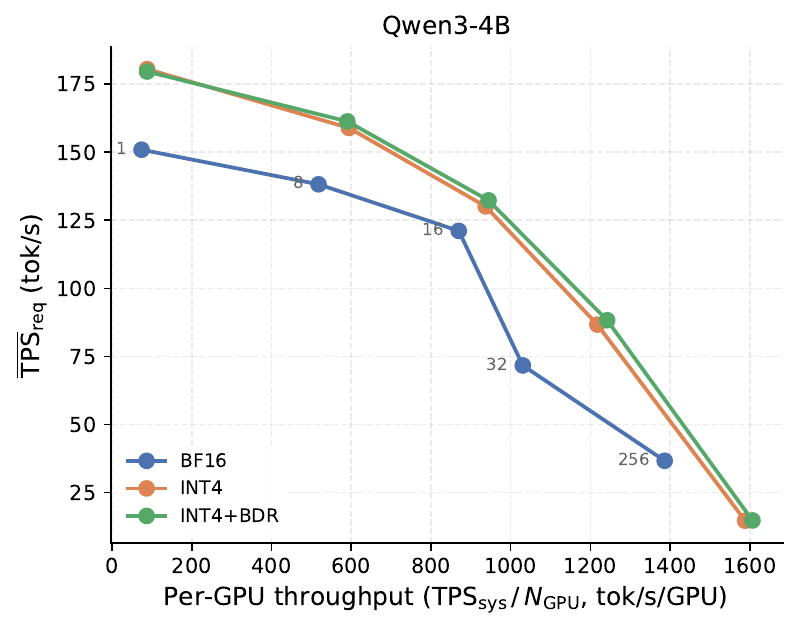}
    \end{minipage}
    \hfill
    \begin{minipage}[t]{0.48\textwidth}
        \centering
        \includegraphics[width=\linewidth]{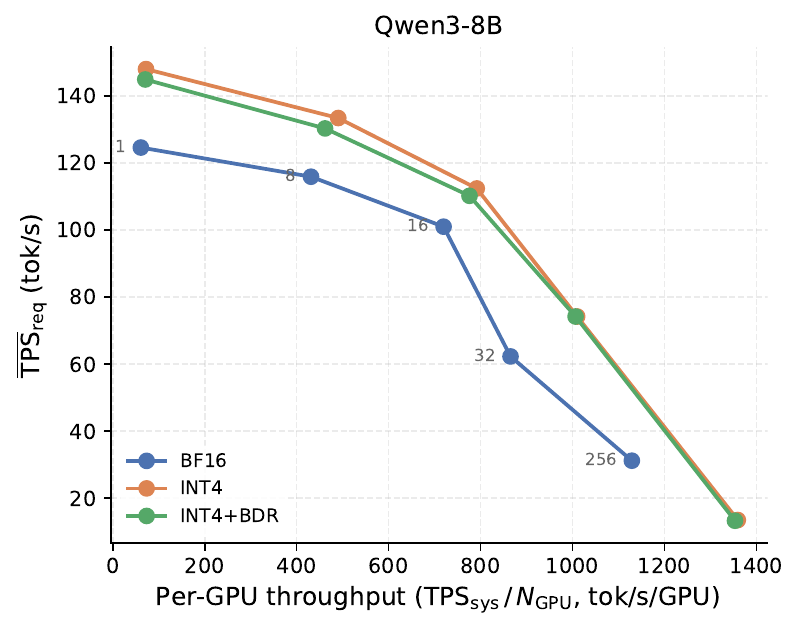}
    \end{minipage}


    \begin{minipage}[t]{0.48\textwidth}
        \centering
       \includegraphics[width=\linewidth]{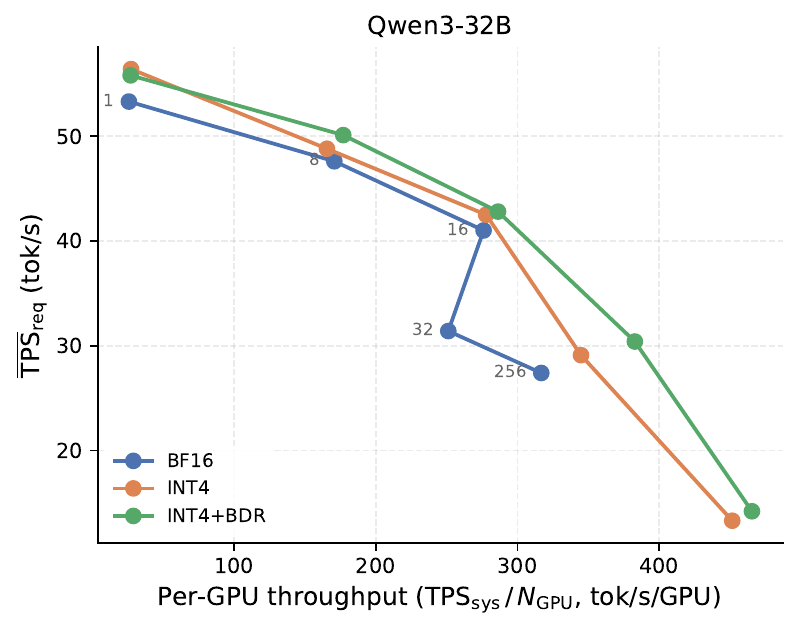}
    \end{minipage}
    \hfill
    \begin{minipage}[t]{0.48\textwidth}
        \centering
        \includegraphics[width=\linewidth]{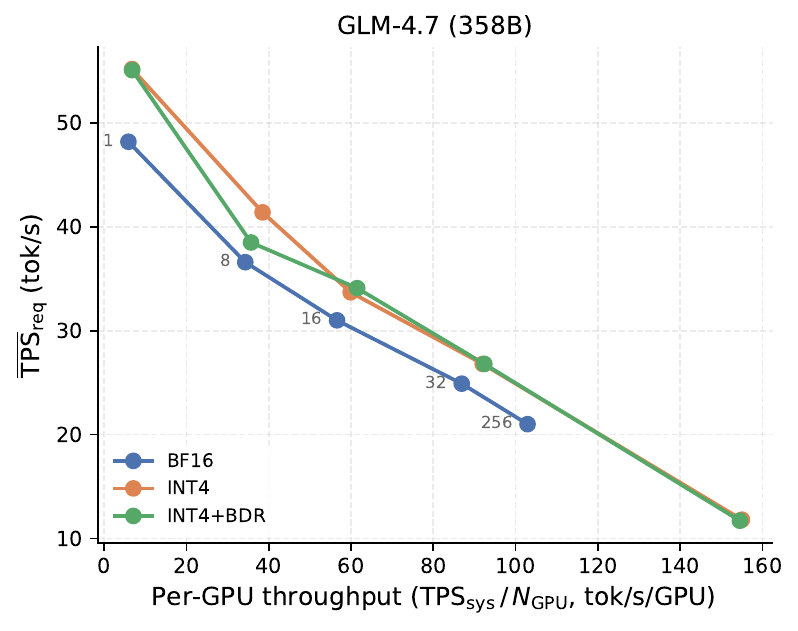}
    \end{minipage}

    \caption{$\overline{\mathrm{TPS}}_\mathrm{req}$ versus per-GPU throughput ($\mathrm{TPS}_\mathrm{sys}/N_\mathrm{GPU}$). The top row shows Qwen3-4B and Qwen3-8B; the bottom row shows Qwen3-32B and GLM-4.7 (358B). INT4 + R128 consistently matches or slightly exceeds plain INT4 and outperforms BF16 across most operating regimes, showing that rotation preserves INT4 efficiency while improving model quality.}
    \label{fig:throughput_2x2}
    \vspace{-1em}
\end{figure*}

This interpretation is consistent with the serving results in Figure~\ref{fig:throughput_2x2}. Across all four workloads, spanning model sizes from Qwen3-4B and Qwen3-8B to Qwen3-32B and GLM-4.7 (358B), INT4 with fused BDR (green curves) consistently matches plain INT4 while outperforming BF16 across most operating regimes. This shows that the small kernel-level advantage of plain INT4 does not translate into a meaningful end-to-end gap. 

We further compare end-to-end throughput against k-means and Hessian-based methods and also include additional comparisons between our INT4 method and Kitty. The complete results are shown in Figure~\ref*{fig:throughput_2x21} and Figure~\ref*{fig:comparison-all} in Appendix~\ref*{sec:ablation_throughput}. Overall, rotation preserves the benefits of INT4 while substantially improving model quality, highlighting that end-to-end performance matters more than isolated kernel microbenchmarks.

\vspace{-0.2cm}

\section{Conclusion}

\vspace{-0.2cm}

This paper asks a simple question: under real serving constraints, what is the minimal INT4 KV-cache quantization method that actually works? Our answer is equally simple: token-wise quantization plus block-diagonal Hadamard rotation. Across models and workloads, this design recovers most of the accuracy lost by naive INT4 while preserving essentially the same serving behavior. More complex alternatives, such as Hessian-aware or vector-quantized variants, add calibration or implementation complexity but yield only marginal gains once serving compatibility is taken into account.

Our results also highlight a broader systems lesson. Kernel microbenchmarks alone can be misleading: a local kernel advantage does not necessarily translate into a meaningful end-to-end serving gain. What matters in practice is preserving the memory-capacity and throughput benefits of INT4 without disrupting the serving stack. By fusing rotation into the KV-cache path, we achieve exactly this, showing that effective KV-cache compression is ultimately a systems co-design problem rather than a purely quantization-driven one.



%% file: section6.tex
\newpage
\section{Details Description of KV Cache Quantization Methods}
\label{sec:appendix-methods}



\subsection{Token-Wise INT4 Quantization}
Token-wise INT4 quantization maps each floating-point key/value activation to 4-bit integers at token granularity. For each token $t$ and attention head $h$, we quantize the head vector $x_{t,h}\in \mathbb{R}^{d_h}$ along the head dimension using one shared scale and zero-point (asymmetric INT4):
\begin{equation}
s_{t,h} = \frac{\max(x_{t,h})-\min(x_{t,h})}{2^4-1}, \qquad
z_{t,h} = \mathrm{round}\!\left(-\frac{\min(x_{t,h})}{s_{t,h}}\right).
\end{equation}
Each element is quantized as
\begin{equation}
q_{t,h,i}=\mathrm{clip}\!\left(\mathrm{round}\!\left(\frac{x_{t,h,i}}{s_{t,h}}\right)+z_{t,h},\,0,\,15\right), \qquad
\hat{x}_{t,h,i}=s_{t,h}(q_{t,h,i}-z_{t,h}).
\end{equation}
In implementation, we pack two neighboring INT4 values into one UINT8 element for compact KV-cache storage and efficient memory transfer.
This token/head-wise design is system-friendly because it preserves tensor layout and reduces KV-cache size by $4\times$ compared to BF16. However, the main source of accuracy degradation is channel-wise range heterogeneity: different channels can have very different value ranges, and this effect is particularly strong for keys. As a result, naive INT4 still introduces large quantization error on sensitive channels and can lead to severe accuracy drops (even near-zero scores in our setting).

\subsection{Vector Quantization (KMeans)}
We evaluate vector quantization based on KMeans, inspired by~\cite{xi2026quantvideogen}. Given a cluster size $C=n$, we first run an offline calibration stage to learn $n$ centroids (codebook entries) for calibrated key/value tokens. During online serving, each newly generated $k,v$ vector is assigned to its nearest centroid, and we store only the centroid index plus a quantized residual.

Formally, for a key vector $k$, let $\mathcal{C}_K=\{c_j\}_{j=1}^{n}$ denote the learned key codebook. We compute
\begin{equation}
\mathrm{id}(k)=\arg\min_{j}\|k-c_j\|_2^2, \qquad
r_k = k-c_{\mathrm{id}(k)}, \qquad
\tilde{r}_k = \mathrm{Quant}(r_k),
\end{equation}
and reconstruct by
\begin{equation}
\hat{k}=c_{\mathrm{id}(k)}+\mathrm{Dequant}(\tilde{r}_k).
\end{equation}
The value branch is processed in the same way using its own codebook $\mathcal{C}_V$. This residual-based coding reduces quantization error compared with naive INT4 because centroid subtraction removes a large low-frequency component before quantization.

Empirically, accuracy improves as $C$ increases, but only up to a saturation point; after that, gains become marginal or unstable depending on the task. We provide detailed results in Table~\ref{tab:centroid_ablation}, where ``$C=n$'' denotes the number of clusters.

\subsection{Hessian-Aware Quantization}
Following and inspired by prior Hessian/covariance-aware quantization approaches~\cite{kim2026happi,zhou2026care}, we derive a learned per-layer orthogonal rotation from offline decode-time query statistics and use it to precondition KV-cache quantization. Let $\ell$ denote a layer, let $q_i \in \mathbb{R}^{d_h}$ be a calibration query vector drawn from that layer, and let $k \in \mathbb{R}^{d_h}$ be a key vector.

\paragraph{Offline calibration.}
We run a calibration set through the serving engine, collect query vectors for each layer, and pool all query heads into a single sample set $\{q_i\}_{i=1}^{N_\ell}$. We then compute the uncentered second-moment matrix
\begin{equation}
M_\ell = \frac{1}{N_\ell}\sum_{i=1}^{N_\ell} q_i q_i^\top \in \mathbb{R}^{d_h \times d_h}.
\end{equation}
Because attention depends on a key only through the score $q^\top k$, the distortion induced by key quantization is weighted by the query distribution, yielding the objective
\begin{equation}
\mathcal{L}_\ell = \delta k^\top M_\ell \delta k.
\end{equation}

\paragraph{Damped eigendecomposition.}
To obtain a stable learned basis, we add diagonal damping before eigendecomposition:
\begin{equation}
\tilde{M}_\ell = M_\ell + \alpha \,\overline{\mathrm{diag}(M_\ell)}\, I,
\end{equation}
where $\alpha = 0.01$ and $\overline{\mathrm{diag}(M_\ell)}$ denotes the mean of the diagonal entries of $M_\ell$. We then compute
\begin{equation}
\tilde{M}_\ell = R_\ell \Lambda_\ell R_\ell^\top,
\end{equation}
and take the orthogonal eigenvector matrix $R_\ell$ as the learned per-layer rotation. In implementation, the eigenvectors are ordered by descending eigenvalue and their signs are canonicalized for reproducibility. Large eigenvalues correspond to attention-sensitive directions, so this basis exposes subspaces where quantization noise is expensive versus negligible.

\paragraph{Runtime application.}
At inference time, we compose $R_\ell$ with a (block) Hadamard transform $H_{(b)}$ and apply the same orthogonal transform to both queries and keys:
\begin{equation}
\hat{q} = (q H_b) R_\ell, \quad \hat{k} = (k H_b) R_\ell.
\end{equation}
Since $H_b R_\ell$ is orthogonal, full-precision attention logits are preserved exactly, i.e., $\hat{q}^\top \hat{k} = q^\top k$. The Hadamard component equalizes within-block magnitudes, while the learned rotation redistributes energy across the full head dimension, jointly reducing the effective dynamic range seen by asymmetric INT4 quantization. Compared with a fixed Hadamard transform alone, this data-driven rotation better matches the layer-wise query distribution, while the extra offline calibration pass is the main practical cost. In our experiments, we instantiate this Hessian-derived rotation with the same block settings as the rest of the paper, including R16 and R128.

For the Value Hessian computation, we use the same calibration pipeline, learned rotation and apply the transform to $V$.

Overall, Hessian-aware calibration learns a layer-specific orthogonal basis from offline query statistics and combines it with a Hadamard transform to steer quantization noise away from attention-sensitive directions while exactly preserving full-precision attention logits. Compared with a fixed rotation, this data-driven construction better matches the layer-wise query distribution, at the cost of an additional offline calibration pass.

As shown in Table~\ref{tab:hwq}, Hessian-aware quantization achieves competitive overall performance on Qwen3-8B comparing to other methods. 

\begin{table}[h!]
\centering
\footnotesize
\setlength{\tabcolsep}{3pt}
\caption{Hessian-Aware Quantization results of Qwen3-8B. Entries are mean scores over 5 runs. Mean is computed over the five task scores.}
\label{tab:hwq}
\resizebox{\columnwidth}{!}{%
\begin{tabular}{l c c c c c c}
\toprule
\textbf{Methods} & \textbf{GPQA} & \textbf{HumanE} & \textbf{LCB v6} & \textbf{AIME} & \textbf{MATH} & \textbf{Mean} \\
\midrule
Hessian+BDR-128 (K only) & 61.23$\pm$2.21 & 87.68$\pm$0.94 & 42.69$\pm$1.77 & 70.00$\pm$4.63 & 89.20$\pm$0.22 & 70.16 \\
Hessian+BDR-128 & 64.71$\pm$1.91 & 86.34$\pm$0.42 & 42.69$\pm$2.40 & 70.00$\pm$5.12 & 89.20$\pm$0.18 & 70.59 \\
\bottomrule
\end{tabular}%
}
\end{table}

\subsection{Block Diagonal Rotation Quantization for Outlier Removal}
The Hadamard transform is a specific type of generalized Fourier transform. We denote the per-head transform as $H_{\mathrm{transform}} \in \mathbb{R}^{d_h \times d_h}$, which satisfies $H_{\mathrm{transform}} = H_{\mathrm{transform}}^\top$ and $H_{\mathrm{transform}} \cdot H_{\mathrm{transform}}^\top = I$ (up to normalization). These properties spread channel outlier information across nearby elements, effectively smoothing extreme activations before quantization while preserving recoverability.

Formally, we apply Hadamard rotation along the head dimension $d_h$:
\begin{equation}
\tilde{K}_{rot} = \mathrm{quant}(K \cdot H_{transform}), \tilde{V}_{rot} = \mathrm{quant}(V \cdot H_{transform}).
\end{equation}
where $H_{\mathrm{transform}}$ is the Hadamard matrix applied along the head dimension.

During decoding, we reconstruct attention using the inverse transform (for Hadamard, $H_{\mathrm{transform}}^{-1}=H_{\mathrm{transform}}^\top$ up to scaling):
\begin{equation}
\mathrm{Score} = \mathrm{softmax}\!\left(\frac{q_{rot} \cdot \mathrm{dequant}(\tilde{K}_{rot})^\top}{\sqrt{d_h}}\right), \quad q_{rot}=q\cdot H_{\mathrm{transform}},
\end{equation}
\begin{equation}
\mathrm{Output} = \mathrm{Score} \cdot \mathrm{dequant}(\tilde{V}_{rot}) \cdot  H_{\mathrm{transform}}.
\end{equation}

Because $H_{\mathrm{transform}}$ is orthogonal, QK products are preserved: $(qH_{\mathrm{transform}})(KH_{\mathrm{transform}})^\top = qH_{\mathrm{transform}}H_{\mathrm{transform}}^\top K^\top = qK^\top$, i.e., $q_{rot}K_{rot}^\top = qK^\top$.

This transformation suppresses outlier effects and substantially improves INT4 quantization fidelity.

Building on the rotation-based formulation in Section~\ref{sec:3}, we propose Block Diagonal Rotation Quantization, which applies a block-diagonal Hadamard transform to improve efficiency while preserving the outlier-mitigation benefit of rotation.

Instead of applying a single transform over the full head dimension, Block Diagonal Rotation partitions the head dimension into smaller chunks and applies independent transforms within each chunk.

This decomposes one large dense rotation into multiple smaller parallel rotations over local feature groups, controlled by the chosen rotation order. The design reduces compute and improves kernel efficiency while retaining sufficient mixing to suppress outliers.

Let $d_h$ denote the head dimension. For each head, we partition key and value vectors into contiguous chunks of size $h$, where $h$ is the given Hadamard order and $h \mid d_h$:
\begin{equation}
K = [K^{(1)}, K^{(2)}, \dots, K^{(d_h/h)}], \quad
V = [V^{(1)}, V^{(2)}, \dots, V^{(d_h/h)}],
\end{equation}
where each chunk lies in $\mathbb{R}^{h}$.

Each chunk is independently transformed and quantized:
\begin{equation}
K_h^{(i)} = \mathrm{quant}(K^{(i)} H_h), \quad
V_h^{(i)} = \mathrm{quant}(V^{(i)} H_h),
\end{equation}
where $H_h \in \mathbb{R}^{h \times h}$ is a Hadamard matrix of order $h$.

This block-diagonal structure gives a practical trade-off between quantization quality and system efficiency, and is straightforward to integrate into serving kernels.

\section{Kernel Profiling Results Across Methods}
\label{sec:appendix-kernel-profiling}
All kernel profiling results in this subsection are measured with \texttt{Qwen3-32B} served by \texttt{SGLang} on two NVIDIA H100 GPUs using tensor parallelism with \texttt{tp=2}. The reported measurements correspond to a single layer.

\begin{table*}[h]
\centering
\scriptsize
\setlength{\tabcolsep}{3.8pt}
\caption{Prefilling kernel profiling results (time in $\mu$s) by category across different input lengths.}
\label{tab:prefill_profiling_detailed}
\begin{tabular}{ll c c c c c}
\toprule
\textbf{Method} & \textbf{Category} & \textbf{8k} & \textbf{16k} & \textbf{32k} & \textbf{64k} & \textbf{128k} \\
\midrule
\multirow{7}{*}{INT4} & MatMul & 5578.24 & 10716.10 & 22077.48 & 42539.09 & 85058.20 \\
 & Attention & 884.96 & 3375.79 & 13516.01 & 51811.20 & 205534.24 \\
 & Communication & 635.80 & 1212.82 & 2319.99 & 4425.88 & 13882.47 \\
 & \_quantized\_set\_kv\_int4\_kernel & 22.05 & 39.04 & 74.11 & 135.87 & 253.47 \\
 & flatten\_dequant & 7.65 & 15.87 & 33.18 & 69.22 & 137.34 \\
 & Others & 364.13 & 719.00 & 1423.26 & 5018.93 & 9942.03 \\
 & TOTAL & 7492.82 & 16078.62 & 39444.04 & 104000.18 & 314807.76 \\
\midrule
\multirow{7}{*}{INT4-Fused-RotateK} & MatMul & 5325.41 & 10313.74 & 21290.69 & 46586.66 & 85555.96 \\
 & Attention & 888.57 & 2933.83 & 13512.67 & 58256.64 & 209316.59 \\
 & Communication & 634.14 & 1593.11 & 2307.48 & 4640.34 & 8361.56 \\
 & \_quantized\_set\_kv\_int4\_kernel & 29.44 & 52.13 & 98.18 & 212.32 & 402.33 \\
 & flatten\_dequant & 21.63 & 38.69 & 75.39 & 165.47 & 313.79 \\
 & Others & 536.16 & 1041.79 & 2076.44 & 5131.22 & 10106.90 \\
 & TOTAL & 7435.35 & 15973.28 & 39360.84 & 114992.65 & 314057.14 \\
\midrule
\multirow{8}{*}{INT4-Unfused-RotateK} & MatMul & 5175.36 & 10536.20 & 21364.65 & 45029.60 & 86323.65 \\
 & Attention & 864.35 & 3315.28 & 13093.66 & 53120.87 & 204113.76 \\
 & Communication & 628.70 & 1202.71 & 2313.11 & 4547.15 & 8508.53 \\
 & \_quantized\_set\_kv\_int4\_kernel & 20.77 & 35.87 & 72.22 & 154.94 & 268.29 \\
 & flatten\_dequant & 7.46 & 14.88 & 32.54 & 72.19 & 140.74 \\
 & Rotation\_unfused\_overhead & 225.63 & 450.85 & 931.29 & 2126.10 & 3864.95 \\
 & Others & 654.17 & 1299.45 & 2579.98 & 5269.13 & 10274.61 \\
 & TOTAL & 7576.44 & 16855.24 & 40387.46 & 110319.99 & 313494.53 \\
\midrule
\multirow{8}{*}{INT4-KMeans-c256} & MatMul & 4952.41 & 11085.02 & 21143.26 & 43554.03 & 87289.12 \\
 & Attention & 821.82 & 3485.32 & 13226.94 & 50640.82 & 204177.53 \\
 & Communication & 619.00 & 1222.59 & 2297.49 & 5095.76 & 8510.50 \\
 & \_quantized\_set\_kv\_int4\_kernel & 72.22 & 163.17 & 306.17 & 676.77 & 1281.66 \\
 & flatten\_dequant & 8.19 & 16.32 & 33.92 & 74.14 & 143.52 \\
 & K-means & 457.95 & 531.93 & 607.14 & 804.93 & 1204.92 \\
 & Others & 746.11 & 1457.63 & 2818.16 & 5631.24 & 11043.55 \\
 & TOTAL & 7677.72 & 17961.98 & 40433.09 & 106477.68 & 313650.80 \\
\bottomrule
\end{tabular}
\end{table*}

\begin{table*}[t]
\centering
\scriptsize
\setlength{\tabcolsep}{3.8pt}
\caption{Decoding kernel profiling results (time in $\mu$s) by category across different batch sizes (input length = 16k).}
\label{tab:decoding_profiling_detailed}
\begin{tabular}{ll c c c c c}
\toprule
\textbf{Method} & \textbf{Category} & \textbf{bs1} & \textbf{bs4} & \textbf{bs8} & \textbf{bs16} & \textbf{bs32} \\
\midrule
\multirow{6}{*}{INT4} & MatMul & 166.43 & 170.46 & 172.03 & 172.77 & 176.80 \\
 & Attention & 130.56 & 138.34 & 135.46 & 191.94 & 326.84 \\
 & Comm & 9.28 & 8.77 & 9.41 & 10.02 & 11.20 \\
 & \_quantized\_set\_kv\_int4\_kernel & 4.61 & 5.18 & 5.50 & 5.86 & 5.50 \\
 & Norm\_Act\_RoPE & 12.00 & 12.71 & 12.03 & 11.52 & 12.80 \\
 & TOTAL & 322.88 & 335.46 & 334.43 & 392.10 & 533.15 \\
\midrule
\multirow{6}{*}{INT4-Fused-RotateK} & MatMul & 169.38 & 169.79 & 171.62 & 173.57 & 176.70 \\
 & Attention & 150.88 & 136.25 & 152.19 & 171.07 & 322.88 \\
 & Comm & 8.54 & 9.06 & 9.73 & 9.28 & 11.52 \\
 & \_quantized\_set\_kv\_int4\_kernel & 6.40 & 6.05 & 6.69 & 6.30 & 6.11 \\
 & Norm\_Act\_RoPE & 13.34 & 13.47 & 13.60 & 12.38 & 12.70 \\
 & TOTAL & 348.54 & 334.62 & 353.82 & 372.61 & 529.92 \\
\midrule
\multirow{7}{*}{INT4-Unfused-RotateK} & MatMul & 168.06 & 170.27 & 175.71 & 173.76 & 175.39 \\
 & Attention & 132.32 & 131.33 & 169.89 & 173.12 & 322.82 \\
 & Comm & 8.19 & 16.16 & 9.57 & 12.64 & 14.50 \\
 & \_quantized\_set\_kv\_int4\_kernel & 4.96 & 5.09 & 6.69 & 5.63 & 5.50 \\
 & Norm\_Act\_RoPE & 12.38 & 12.35 & 14.53 & 12.48 & 13.03 \\
 & Rotation\_unfused\_overhead & 4.96 & 8.32 & 10.53 & 8.93 & 9.38 \\
 & TOTAL & 330.88 & 343.52 & 386.91 & 386.56 & 540.61 \\
\midrule
\multirow{7}{*}{INT4-KMeans-c256} & MatMul & 177.22 & 176.64 & 176.77 & 178.59 & 181.41 \\
 & Attention & 180.67 & 202.89 & 208.10 & 600.16 & 834.24 \\
 & Comm & 11.72 & 9.12 & 20.48 & 9.15 & 94.67 \\
 & \_quantized\_set\_kv\_int4\_kernel & 3.14 & 3.14 & 3.20 & 3.39 & 3.39 \\
 & Norm\_Act\_RoPE & 12.23 & 12.45 & 12.38 & 12.67 & 12.91 \\
 & KMeans\_Overhead & 132.66 & 234.93 & 256.35 & 264.77 & 297.25 \\
 & TOTAL & 517.63 & 639.16 & 677.28 & 1068.74 & 1423.88 \\
\bottomrule
\end{tabular}
\begin{flushleft}
{\scriptsize
\textit{Notes.}
MatMul: GEMM kernels; 
Attention: decode-attention kernels; 
Comm: \texttt{cross\_device\_reduce\_*}; 
\texttt{\_quantized\_set\_kv\_int4\_kernel}: INT4 KV-cache write (including fused Hadamard + KV write); 
Norm\_Act\_RoPE: RMSNorm, activation, RoPE, and QK-norm; 
Outside\_Kernel\_Rotation: unfused rotation overhead; 
KMeans: clustering overhead, including Euclidean distance computation and centroid assignment.
}
\end{flushleft}
\end{table*}

Table~\ref{tab:prefill_profiling_detailed} presents the detailed kernel-level overhead during the prefill phase, evaluated across input lengths ranging from 8K to 128K tokens. For the unfused rotation implementation, the overhead decreases from approximately 3\% to 1.2\% as the input length increases, since other components dominate the runtime at longer sequence lengths. In contrast, the K-means-based method introduces an additional overhead of 6\% to 0.3\% compared to pure INT4, primarily due to the cost of cluster assignment and centroid computation for each token.

Although the fused rotation kernel incurs a 1.33$\times$ to 1.58$\times$ increase in the execution time of the \texttt{quantized_set_kv} and \texttt{flatten_dequant} kernels, this overhead contributes only 0.28\% to 0.10\% to the overall runtime, making it negligible in practice.

Table~\ref{tab:decoding_profiling_detailed} reports the kernel profiling results for the decoding phase. The fused rotation approach shows comparable overhead to pure INT4, as the KV quantization cost is minimal due to the small per-token workload and latency-dominated execution. In contrast, the unfused rotation introduces an additional overhead of around 1.5\%.

Despite leveraging a Flash-KMeans kernel \cite{flashkmeans2026} for centroid computation and fusing the residual addition into the attention kernel, the K-means method still incurs significant overhead during decoding. This overhead mainly arises from per-token clustering for newly generated tokens, as well as additional computation within the attention kernel. Notably, the attention kernel under K-means is 1.4$\times$ to 2.5$\times$ slower than that of pure INT4, which we attribute to the additional memory loading overhead for centroids introduced in the attention kernel.

\clearpage
\section{Ablation of Token-Wise Quantization}
\label{sec:ablation}

This section ablates two design choices in our token-wise INT4 pipeline: (i) the rotation order and target (keys only vs.\ keys and values) for block-diagonal Hadamard rotation, and (ii) the codebook size for residual vector quantization on top of BDR.

\paragraph{Table~\ref{tab:kv_compare1}: rotation order and target.}
We sweep BDR rotation orders $h \in \{16, 64, 128\}$ with rotation applied to keys only (K) or to both keys and values (K\&V).
Two conclusions emerge.
First, \emph{rotating keys only is sufficient}: BDR(K) and BDR(K\&V) achieve nearly identical mean accuracy across all three models and all rotation orders, so adding value rotation yields negligible benefit and we default to key-only rotation to reduce the runtime overhead.
Second, \emph{higher rotation orders are critical for smaller, more quantization-sensitive models}: for Qwen3-4B, BDR16 incurs a mean drop of up to $-22.5$ points relative to BF16, while BDR128 reduces the gap to $-1.9$ points.
For Qwen3-8B and GLM-4.7-FP8 the sensitivity to order is smaller, though BDR64 and BDR128 still consistently outperform BDR16.

\paragraph{Table~\ref{tab:centroid_ablation}: centroid size for residual coding.}
We calibrate KV-cache centroids using MMLU-Pro (first 20{,}000 tokens) and test codebook sizes $C \in \{1, 16, 256, 2048\}$ combined with BDR at orders 16, 64, and 128.
Note that KMeans centroid coding is applied to \emph{both} keys and values in all settings, while the ``(K)'' suffix refers only to the BDR rotation target (keys only).
For Qwen3-4B, residual coding with $C{=}256$ yields consistent gains over pure BDR+INT4 ($C{=}1$), recovering roughly $1$--$2$ additional mean points depending on the rotation order.
However, scaling to $C{=}2048$ provides no further improvement and in some settings slightly regresses, indicating accuracy saturates around $C{=}256$.
For Qwen3-8B the effect of centroid size is small and inconsistent across orders, suggesting BDR alone already captures most of the quantization error structure at that model scale.
Given the additional calibration and memory overhead of large codebooks, $C{=}256$ represents a practical sweet spot.

\begin{table*}[h]
\centering
\scriptsize
\setlength{\tabcolsep}{3.8pt}
\caption{Benchmark results of different KV cache configurations. Entries are $\mu \pm \sigma$.}
\label{tab:kv_compare1} 
\begin{tabular}{l c c c c c c c}
\toprule
\textbf{Method}
& \makecell{\textbf{GPQA}\\\textbf{Diamond}}
& \makecell{\textbf{Human}\\\textbf{Eval}}
& \makecell{\textbf{LiveCode}\\\textbf{Bench (v6)}}
& \textbf{AIME25}
& \makecell{\textbf{MATH}\\\textbf{500}}
& \textbf{Mean}
& \textbf{Drop} \\
\midrule

\multicolumn{8}{c}{\textbf{Qwen3-4B-Thinking-2507}} \\
\midrule
BF16                   & 67.27$\pm$1.80 & 94.05$\pm$0.54 & 48.66$\pm$2.20 & 74.67$\pm$1.83 & 93.55$\pm$0.33 & 75.64 & -- \\
INT4 (K\&V)            & 0.00$\pm$0.00 & 0.00$\pm$0.00 & 0.00$\pm$0.00 & 0.00$\pm$0.00 & 0.00$\pm$0.00 & 0.00 & -- \\
\midrule
BDR16(K)        & 51.72$\pm$2.22 & 57.12$\pm$1.91 & 21.99$\pm$1.97 & 48.00$\pm$1.82 & 86.65$\pm$1.56 & 53.10 & -22.54 \\
BDR16(K\&V)     & 50.61$\pm$2.35 & 56.20$\pm$2.40 & 26.20$\pm$2.85 & 54.67$\pm$4.47 & 86.45$\pm$0.61 & 54.83 & -20.81 \\
BDR64(K)        & 63.33$\pm$1.36 & 88.56$\pm$0.95 & 46.20$\pm$1.71 & 69.33$\pm$4.35 & 93.19$\pm$0.32 & 72.12 & -3.52 \\
BDR64(K\&V)     & 63.13$\pm$1.38 & 89.10$\pm$0.39 & 46.66$\pm$1.33 & 69.34$\pm$4.35 & 93.23$\pm$0.52 & 72.29 & -3.35 \\
BDR128(K)       & 65.25$\pm$2.12 & 90.49$\pm$0.45 & 48.54$\pm$1.65 & 71.33$\pm$5.06 & 93.27$\pm$0.36 & 73.78 & -1.86 \\
BDR128(K\&V)    & 66.37$\pm$2.19 & 89.78$\pm$0.80 & 46.20$\pm$1.94 & 70.00$\pm$4.08 & 93.19$\pm$0.51 & 73.11 & -2.53 \\

\midrule
\multicolumn{8}{c}{\textbf{Qwen3-8B}} \\
\midrule
BF16                   & 56.67$\pm$2.30 & 85.95$\pm$1.01 & 49.01$\pm$2.13 & 70.00$\pm$3.33 & 92.59$\pm$0.62 & 70.84 & -- \\
INT4 (K\&V)            & 0.00$\pm$0.00 & 0.00$\pm$0.00 & 0.00$\pm$0.00 & 0.00$\pm$0.00 & 0.00$\pm$0.00 & 0.00 & -- \\
\midrule
BDR16(K)        & 53.44$\pm$3.45 & 84.66$\pm$0.72 & 44.68$\pm$2.36 & 64.00$\pm$7.96 & 91.90$\pm$0.41 & 67.74 & -3.11 \\
BDR16(K\&V)     & 54.85$\pm$2.44 & 86.05$\pm$0.80 & 47.13$\pm$1.52 & 59.33$\pm$5.96 & 92.06$\pm$0.86 & 67.88 & -2.96 \\
BDR64(K)        & 56.97$\pm$0.83 & 86.59$\pm$0.98 & 47.02$\pm$1.58 & 66.00$\pm$4.94 & 92.71$\pm$0.37 & 69.86 & -0.99 \\
BDR64(K\&V)     & 54.55$\pm$1.55 & 87.59$\pm$0.48 & 47.48$\pm$2.16 & 64.00$\pm$4.35 & 92.18$\pm$0.64 & 69.16 & -1.68 \\
BDR128(K)       & 53.23$\pm$2.59 & 86.81$\pm$0.46 & 46.78$\pm$0.92 & 61.33$\pm$6.06 & 92.59$\pm$0.29 & 68.15 & -2.70 \\
BDR128(K\&V)    & 54.85$\pm$3.17 & 86.44$\pm$0.42 & 47.95$\pm$2.15 & 68.00$\pm$2.98 & 92.63$\pm$0.27 & 69.97 & -0.87 \\

\midrule

\multicolumn{8}{c}{\textbf{GLM-4.7-FP8}} \\
\midrule
BF16                   & 73.23$\pm$1.33 & 91.46$\pm$0.65 & 49.12$\pm$0.59 & 80.00$\pm$3.33 & 95.66$\pm$0.61 & 77.89 & -- \\
INT4 (K\&V)            & 73.23$\pm$1.34 & 91.14$\pm$0.26 & 45.81$\pm$1.79 & 80.00$\pm$0.00 & 95.86$\pm$0.31 & 77.21 & -0.68 \\
\midrule
BDR16(K)        & 72.39$\pm$0.58 & 91.83$\pm$0.53 & 51.46$\pm$3.10 & 81.11$\pm$1.92 & 95.26$\pm$0.31 & 78.41 & +0.52 \\
BDR16(K\&V)     & 72.90$\pm$0.29 & 91.30$\pm$0.63 & 50.68$\pm$0.89 & 78.89$\pm$1.92 & 95.46$\pm$0.42 & 77.85 & -0.04 \\
BDR64(K)        & 74.92$\pm$2.78 & 90.98$\pm$0.53 & 47.56$\pm$3.90 & 76.67$\pm$3.34 & 95.46$\pm$0.23 & 77.12 & -0.77 \\
BDR64(K\&V)     & 70.87$\pm$1.54 & 92.11$\pm$0.31 & 49.90$\pm$1.79 & 81.11$\pm$3.85 & 95.59$\pm$0.35 & 77.92 & +0.03 \\
BDR128(K)       & 73.23$\pm$2.53 & 91.46$\pm$0.13 & 49.12$\pm$3.83 & 77.78$\pm$1.92 & 95.19$\pm$0.40 & 77.36 & -0.53 \\
BDR128(K\&V)    & 73.74$\pm$1.01 & 91.30$\pm$0.92 & 50.49$\pm$1.47 & 78.89$\pm$1.92 & 95.32$\pm$0.12 & 77.95 & +0.06 \\
\bottomrule
\end{tabular}

\vspace{0.3em}
\footnotesize
\begin{flushleft}
\scriptsize
\emph{Notes.} BF16 denotes full-precision KV cache. INT4 denotes uniform 4-bit quantization. 
BDR refers to the Block Diagonal Rotation algorithm. 
Suffixes “(K)” and “(K\&V)” indicate whether rotation is applied to key only or both key and value. 
All experiments use a maximum generation length of 32768 tokens.
\end{flushleft}
\end{table*}

\begin{table*}[h]
\centering
\scriptsize
\setlength{\tabcolsep}{3.8pt}
\caption{Effect of centroid size under different rotation ranks on KV cache quantization (Qwen3-4B-Thinking-2507). Entries are $\mu \pm \sigma$.}
\label{tab:centroid_ablation}
\begin{tabular}{l c c c c c c c}
\toprule
\textbf{Method}
& \makecell{\textbf{GPQA}\\\textbf{Diamond}}
& \makecell{\textbf{Human}\\\textbf{Eval}}
& \makecell{\textbf{LiveCode}\\\textbf{Bench (v6)}}
& \textbf{AIME25}
& \makecell{\textbf{MATH}\\\textbf{500}}
& \textbf{Mean}
& \textbf{Drop} \\
\midrule

\multicolumn{8}{c}{\textbf{Qwen3-4B-Thinking-2507}} \\
\midrule
BF16                      & 67.27$\pm$1.80 & 94.05$\pm$0.54 & 48.66$\pm$2.20 & 74.67$\pm$1.83 & 93.55$\pm$0.33 & 75.64 & -- \\
INT4            & 0.00$\pm$0.00 & 0.00$\pm$0.00 & 0.00$\pm$0.00 & 0.00$\pm$0.00 & 0.00$\pm$0.00 & 0.00 & -- \\

\midrule

C=1 BDR16 (K)    & 64.14$\pm$1.60 & 92.39$\pm$0.61 & 43.27$\pm$2.22 & 66.67$\pm$5.58 & 93.15$\pm$0.32 & 71.92 & -3.72 \\
C=16 BDR16 (K)   & 66.26$\pm$1.44 & 93.49$\pm$0.52 & 44.56$\pm$1.63 & 66.67$\pm$4.71 & 93.55$\pm$0.08 & 72.91 & -2.73 \\
C=256 BDR16 (K)  & 65.86$\pm$2.38 & 93.98$\pm$0.64 & 46.43$\pm$2.08 & 72.00$\pm$3.40 & 94.31$\pm$0.68 & 74.52 & -1.12 \\
C=2048 BDR16 (K) & 64.75$\pm$1.76 & 92.78$\pm$0.81 & 46.55$\pm$1.68 & 71.33$\pm$6.87 & 93.67$\pm$0.27 & 73.82 & -1.82 \\

\midrule

C=1 BDR64 (K)    & 63.84$\pm$2.27 & 93.17$\pm$0.49 & 43.16$\pm$2.04 & 68.67$\pm$4.52 & 93.47$\pm$0.52 & 72.46 & -3.18 \\
C=16 BDR64 (K)   & 66.97$\pm$0.88 & 93.61$\pm$0.69 & 46.08$\pm$1.13 & 70.00$\pm$3.65 & 93.51$\pm$0.27 & 74.03 & -1.61 \\
C=256 BDR64 (K)  & 64.34$\pm$1.93 & 93.32$\pm$0.37 & 44.68$\pm$0.95 & 71.33$\pm$3.40 & 93.51$\pm$0.41 & 73.44 & -2.20 \\
C=2048 BDR64 (K) & 65.56$\pm$1.87 & 94.05$\pm$0.21 & 46.90$\pm$1.94 & 71.33$\pm$7.48 & 93.51$\pm$0.45 & 74.27 & -1.37 \\

\midrule

C=1 BDR128 (K)    & 64.65$\pm$1.01 & 94.49$\pm$0.43 & 43.27$\pm$2.42 & 70.00$\pm$4.22 & 93.35$\pm$0.56 & 73.15 & -2.49 \\
C=16 BDR128 (K)   & 63.33$\pm$1.38 & 93.17$\pm$0.20 & 43.86$\pm$1.57 & 68.67$\pm$3.40 & 93.55$\pm$0.51 & 72.52 & -3.12 \\
C=256 BDR128 (K)  & 64.65$\pm$1.81 & 92.93$\pm$0.66 & 45.73$\pm$1.45 & 72.00$\pm$2.67 & 93.43$\pm$0.77 & 73.75 & -1.89 \\
C=2048 BDR128 (K) & 65.25$\pm$0.49 & 93.27$\pm$0.63 & 47.25$\pm$1.63 & 67.33$\pm$2.50 & 93.63$\pm$0.51 & 73.35 & -2.29 \\
\midrule
\multicolumn{8}{c}{\textbf{Qwen3-8B}} \\
\midrule
BF16                & 56.67$\pm$2.30 & 85.95$\pm$1.01 & 49.01$\pm$2.13 & 70.00$\pm$3.33 & 92.59$\pm$0.62 & 70.84 & -- \\
INT4                & 5.56$\pm$0.00  & 4.39$\pm$0.00  & 0.58$\pm$0.00  & 0.00$\pm$0.00  & 0.00$\pm$0.00  & 2.11  & -68.73 \\
\midrule
C=1 BDR128 (K)    & 57.38$\pm$1.65 & 87.17$\pm$0.46 & 48.30$\pm$1.91 & 67.33$\pm$7.42 & 92.47$\pm$0.59 & 70.53 & -0.31 \\
C=16 BDR128 (K)   & 56.77$\pm$1.13 & 87.46$\pm$0.76 & 48.07$\pm$1.01 & 66.67$\pm$3.65 & 92.38$\pm$0.49 & 70.27 & -0.57 \\
C=256 BDR128 (K)  & 56.16$\pm$2.77 & 88.12$\pm$0.29 & 47.72$\pm$1.83 & 64.67$\pm$5.41 & 92.51$\pm$0.52 & 69.84 & -1.00 \\
C=2048 BDR128 (K) & 57.07$\pm$1.83 & 87.88$\pm$0.56 & 47.49$\pm$2.41 & 64.67$\pm$1.64 & 92.67$\pm$0.45 & 69.96 & -0.88 \\

\bottomrule
\end{tabular}

\vspace{0.3em}
\scriptsize
\emph{Notes.} BF16 denotes full-precision KV cache. INT4 denotes uniform 4-bit quantization.
BDR refers to Block Diagonal Rotation; all BDR settings use INT4 quantization.
$C$ denotes the number of KMeans centroids; centroid-based residual coding is applied to \emph{both} keys and values in all rows.
Suffix ``(K)'' indicates that BDR rotation is applied to keys only.
The notation ``C$=n$ BDR$h$ (K)'' means $n$ KMeans centroids (K\&V) combined with BDR of order $h$ on keys only.
All experiments use a maximum generation length of 32768 tokens.
\end{table*}

%% file: section5.tex
\clearpage
\section{Serving Throughput Under Concurrent Load}
\label{sec:ablation_throughput}

\paragraph{Metrics.}
We report three complementary metrics throughout this section.
\textbf{System TPS} measures aggregate serving efficiency as total output tokens divided by wall-clock time for the entire benchmark:
\begin{equation}
  \mathrm{TPS}_\mathrm{sys} = \frac{\displaystyle\sum_{i=1}^{N} L_i^{\mathrm{out}}}{T_{\mathrm{wall}}},
  \label{eq:sys_tps}
\end{equation}
where $N$ is the total number of completed requests, $L_i^{\mathrm{out}}$ is the output length of request $i$, and $T_{\mathrm{wall}}$ is the elapsed wall-clock time.
\textbf{Per-request output TPS} ($\overline{\mathrm{TPS}}_\mathrm{req}$) averages the decode-phase throughput seen by each individual request:
\begin{equation}
  \overline{\mathrm{TPS}}_\mathrm{req} = \frac{1}{N}\sum_{i=1}^{N} \frac{L_i^{\mathrm{out}}}{t_i^{\mathrm{e2e}} - \mathrm{TTFT}_i},
  \label{eq:req_tps}
\end{equation}
where $t_i^{\mathrm{e2e}}$ is the end-to-end latency of request $i$.
\textbf{Time-to-first-token (TTFT)} measures serving-side responsiveness:
\begin{equation}
  \mathrm{TTFT}_i = t_i^{\mathrm{first\_token}} - t_i^{\mathrm{submit}}.
  \label{eq:ttft}
\end{equation}
Note that $\mathrm{TPS}_\mathrm{sys}$ (Eq.~\ref*{eq:sys_tps}) captures the \emph{server}-side efficiency, while $\overline{\mathrm{TPS}}_\mathrm{req}$ (Eq.~\ref*{eq:req_tps}) reflects the \emph{per-client} decode speed during active generation.
These two metrics can diverge significantly under memory-pressure conditions, as we explain below.

\paragraph{Setup.}
We evaluate on a single NVIDIA H100 80GB GPU with Qwen3-8B under two workload scenarios:
\textbf{long context} (mean input 16{,}384 tokens, 1{,}024 output tokens) and \textbf{short context} (mean input 256 tokens, 1{,}024 output tokens).
We compare BF16 (baseline), INT4 (token-wise KV quantization), and INT4+BDR (INT4 with block-diagonal Hadamard rotation, $\mathrm{ord}{=}128$, keys only) across a sweep of concurrency levels.

\begin{figure}[t]
  \centering
  \begin{minipage}[b]{0.48\linewidth}
    \centering
    \includegraphics[width=\linewidth]{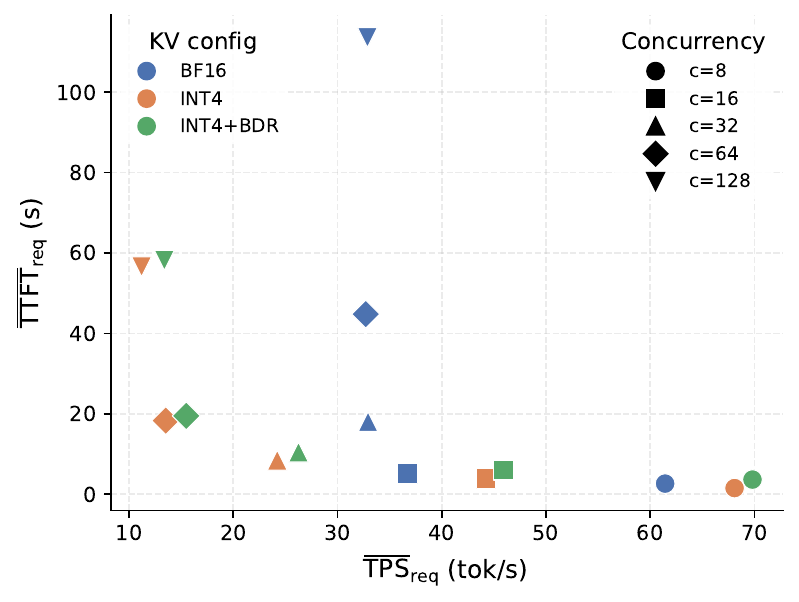}
    \small{(a) Long context: $\overline{\mathrm{TPS}}_\mathrm{req}$ vs.\ $\overline{\mathrm{TTFT}}_\mathrm{req}$.}
  \end{minipage}
  \hfill
  \begin{minipage}[b]{0.48\linewidth}
    \centering
    \includegraphics[width=\linewidth]{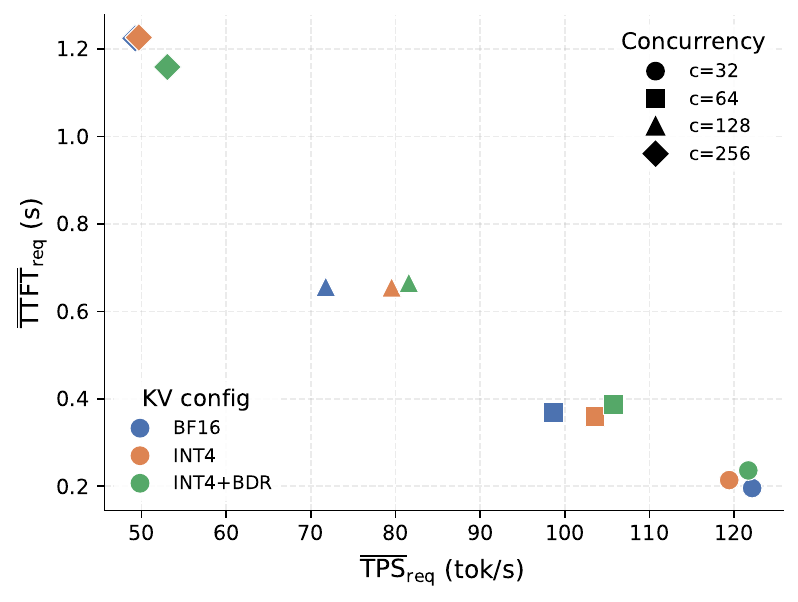}
    \small{(b) Short context: $\overline{\mathrm{TPS}}_\mathrm{req}$ vs.\ $\overline{\mathrm{TTFT}}_\mathrm{req}$.}
  \end{minipage}
  \caption{$\overline{\mathrm{TPS}}_\mathrm{req}$ vs.\ $\overline{\mathrm{TTFT}}_\mathrm{req}$ on 1$\times$H100 (Qwen3-8B). Marker shape encodes concurrency level; color encodes KV cache configuration (blue = BF16, red = INT4, green = INT4+BDR). Points to the right and lower are better. In the long-context regime at high concurrency, BF16 achieves artificially high $\overline{\mathrm{TPS}}_\mathrm{req}$ by running small batches due to its larger KV memory footprint, but pays a severe TTFT penalty; system-level throughput is shown separately.}
  \label{fig:serving_scatter}
\end{figure}

\begin{figure}[t]
  \centering
  \begin{minipage}[b]{0.48\linewidth}
    \centering
    \includegraphics[width=\linewidth]{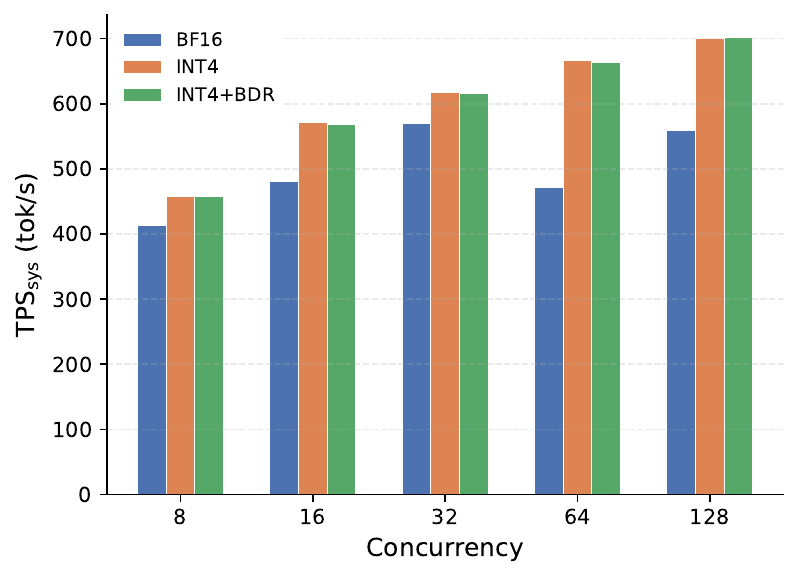}
    \small{(a) Long context: system TPS ($\mathrm{TPS}_\mathrm{sys}$).}
  \end{minipage}
  \hfill
  \begin{minipage}[b]{0.48\linewidth}
    \centering
    \includegraphics[width=\linewidth]{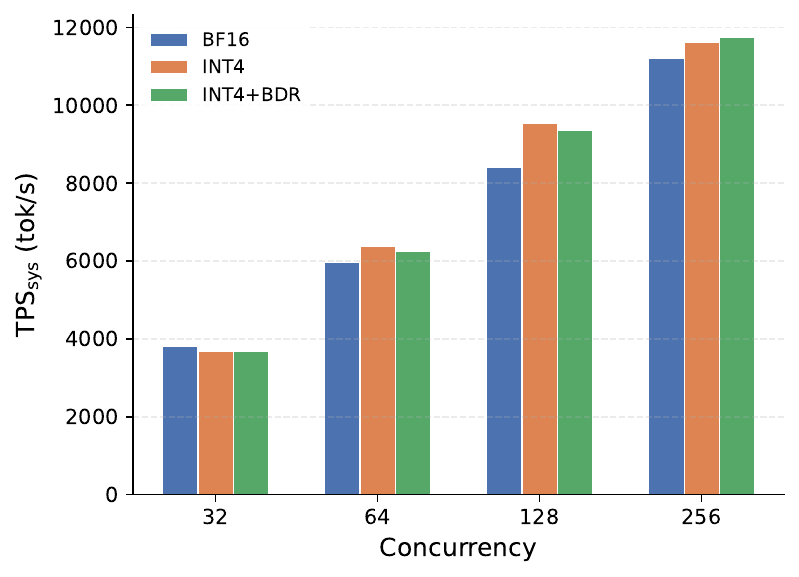}
    \small{(b) Short context: system TPS ($\mathrm{TPS}_\mathrm{sys}$).}
  \end{minipage}
  \caption{System-level throughput ($\mathrm{TPS}_\mathrm{sys}$) on 1$\times$H100 (Qwen3-8B) at varying concurrency levels. In the long-context regime (a), INT4 and INT4+BDR \emph{consistently} outperform BF16 at all concurrency levels by $+8$--$41\%$, resolving the apparent per-request TPS paradox. In the short-context regime (b), INT4 and INT4+BDR are neutral at low concurrency and increasingly advantageous at higher concurrency.}
  \label{fig:serving_sys_tps}
\end{figure}

\paragraph{Short-context regime.}
Figures~\ref*{fig:serving_scatter} and~\ref*{fig:serving_sys_tps}, panel~(b), show results for the short-context workload.
When KV cache memory is not the bottleneck, the three configurations behave similarly.
At low concurrency (32), BF16 is marginally faster by ${\sim}3\%$ on $\mathrm{TPS}_\mathrm{sys}$ and ${\sim}1\%$ on $\overline{\mathrm{TPS}}_\mathrm{req}$, reflecting the small overhead of INT4 quantization when the GPU is underutilized.
At higher concurrency (${\geq}64$), INT4 and INT4+BDR achieve progressively larger system TPS gains---up to $+13.5\%$ and $+11.2\%$ at concurrency 128 respectively---as the reduced KV footprint begins to allow slightly larger decode batches.
TTFT remains comparable across all configurations in this regime.

\paragraph{Long-context regime: a tale of two throughput metrics.}
The long-context results reveal a subtle but important distinction between $\overline{\mathrm{TPS}}_\mathrm{req}$ and $\mathrm{TPS}_\mathrm{sys}$.
Figure~\ref*{fig:serving_scatter}, panel~(a), shows the per-request TPS vs.\ TTFT trade-off.
At low concurrency (${\leq}16$), INT4 and INT4+BDR \emph{both} outperform BF16 on $\overline{\mathrm{TPS}}_\mathrm{req}$ and achieve lower TTFT simultaneously (points to the right \emph{and} below BF16).
At high concurrency (${\geq}32$), however, INT4 and INT4+BDR fall \emph{to the left} of BF16 on the scatter plot, appearing to deliver lower per-request TPS despite still achieving much lower TTFT.

This apparent paradox has a mechanical explanation rooted in KV cache memory capacity.
A BF16 KV cache entry occupies $4{\times}$ the memory of an INT4 entry; at the same GPU memory budget, BF16 can buffer only $0.25{\times}$ as many token slots as INT4.
Under high concurrency, the BF16 engine is therefore forced to run with a \emph{much smaller} effective decode batch size---most requests sit in the queue waiting for a free KV slot.
Smaller batches mean fewer memory-bound attention operations per decode step, which \emph{artificially inflates} the per-request decode speed $\overline{\mathrm{TPS}}_\mathrm{req}$ for BF16.
Meanwhile, queued requests accumulate enormous TTFT (exceeding 44s at concurrency 64 and 113s at concurrency 128 for BF16, vs.\ 19s and 57s for INT4).
The true picture emerges from $\mathrm{TPS}_\mathrm{sys}$: as shown in Figure~\ref*{fig:serving_sys_tps}, panel~(a), INT4 and INT4+BDR achieve \emph{consistently higher system TPS than BF16 at every concurrency level}, with gains of $+10.7\%$ at concurrency 8, $+18.7\%$ at 16, $+8.4\%$ at 32, $+41.4\%$ at 64, and $+25.4\%$ at 128.
INT4+BDR matches INT4 system TPS within ${\leq}1\%$ at all concurrency levels, confirming that the Hadamard rotation overhead is negligible end-to-end.

In summary, $\overline{\mathrm{TPS}}_\mathrm{req}$ can be misleading under memory-pressure conditions because it measures only the active decode phase and ignores queuing delays.
$\mathrm{TPS}_\mathrm{sys}$ (Eq.~\ref*{eq:sys_tps}) is the appropriate metric for comparing serving efficiency across configurations with different KV cache memory footprints.

\begin{figure*}[t]
    \centering
    \begin{minipage}[t]{0.48\textwidth}
        \centering
        \includegraphics[width=\linewidth]{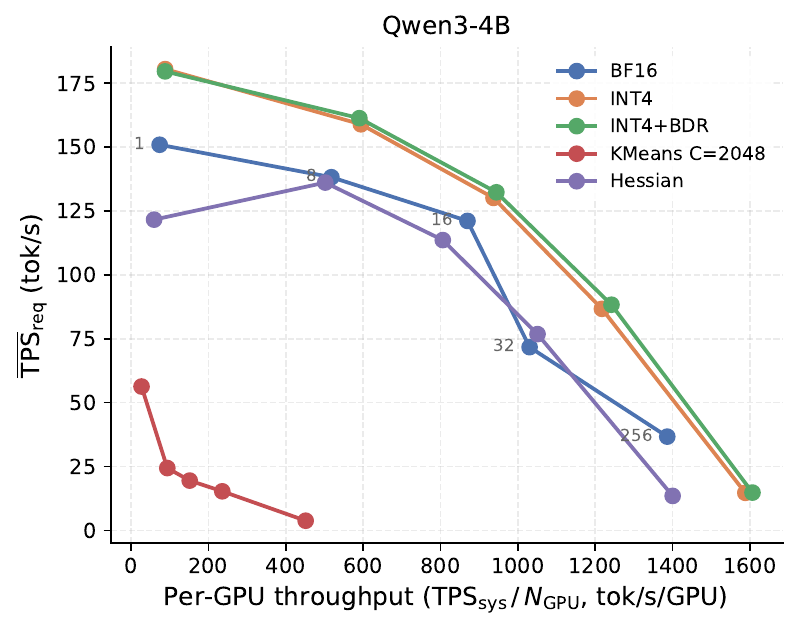}
    \end{minipage}
    \hfill
    \begin{minipage}[t]{0.48\textwidth}
        \centering
        \includegraphics[width=\linewidth]{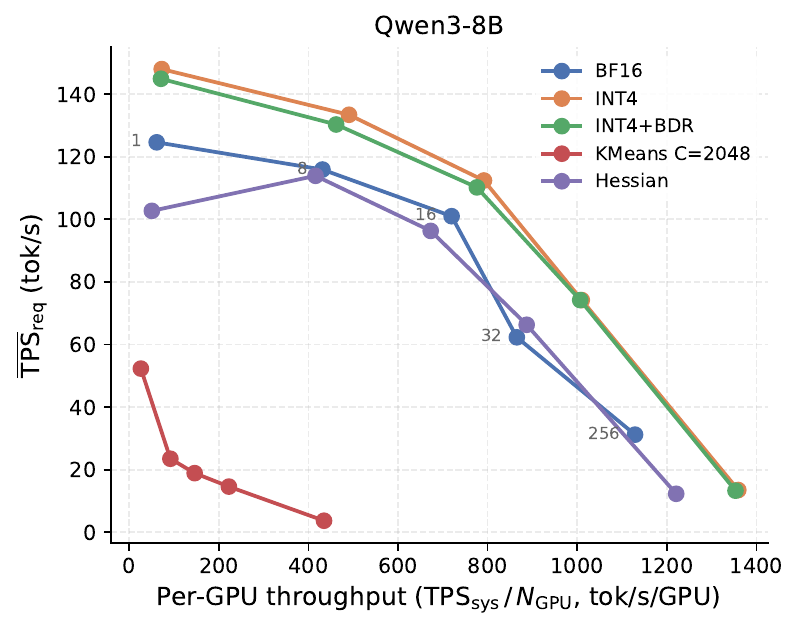}
    \end{minipage}

    \begin{minipage}[t]{0.48\textwidth}
        \centering
       \includegraphics[width=\linewidth]{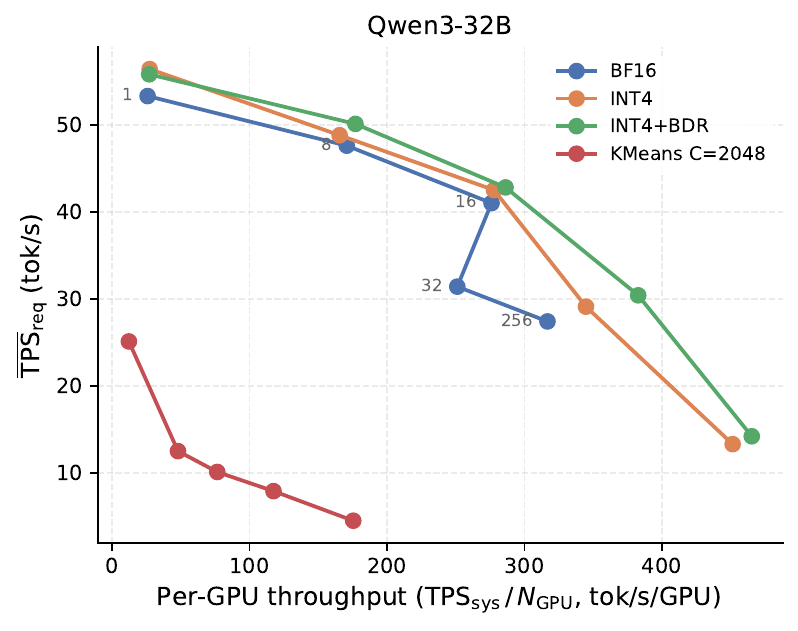}
    \end{minipage}
    \hfill
    \begin{minipage}[t]{0.48\textwidth}
        \centering
        \includegraphics[width=\linewidth]{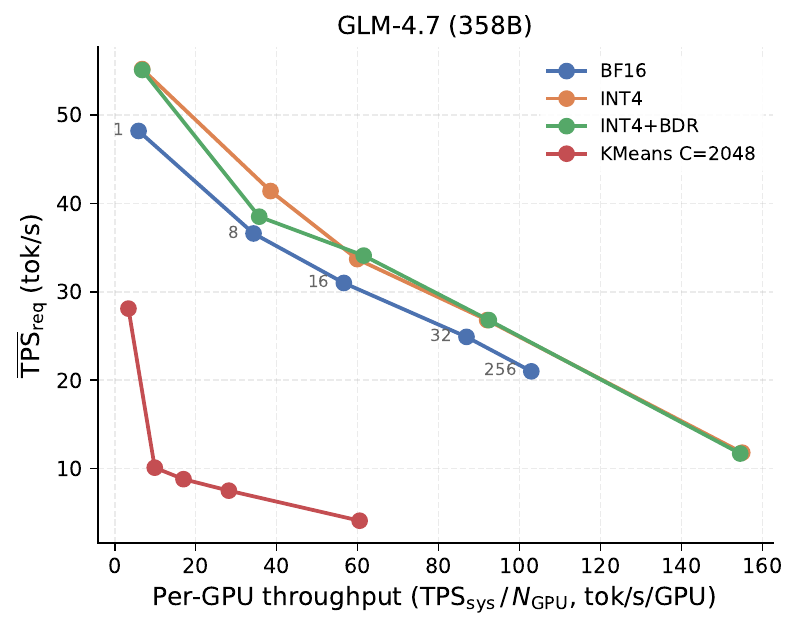}
    \end{minipage}

    \caption{$\overline{\mathrm{TPS}}_\mathrm{req}$ versus per-GPU throughput ($\mathrm{TPS}_\mathrm{sys}/N_\mathrm{GPU}$) across four workloads. The top row shows Qwen3-4B and Qwen3-8B; the bottom row shows Qwen3-32B and GLM-4.7 (358B). INT4 + R128 consistently matches or slightly exceeds plain INT4 and outperforms BF16 across most operating regimes, showing that rotation preserves INT4 efficiency while improving model quality.}
    \label{fig:throughput_2x21}
    \vspace{-1em}
\end{figure*}

\paragraph{Generalization across model sizes and hardware configurations.}
To evaluate whether the efficiency gains hold more broadly, we extend the comparison to four model--hardware pairings:
Qwen3-4B and Qwen3-8B served on 2$\times$H100 80GB with \texttt{tp=2},
Qwen3-32B served on 2$\times$H100 80GB with \texttt{tp=2},
and GLM-4.7-FP8 (358B) served on 8$\times$H100 80GB with \texttt{tp=8}.
All experiments use SGLang with FA3 prefill and Triton decode backends, KV cache memory fraction 0.8, temperature 0.6, top-$p$ 0.95, mean input length $\approx$8{,}192 tokens, and output length $\approx$1{,}024 tokens.
We sweep concurrency from 1 to 256.
Figure~\ref{fig:throughput_2x21} plots mean per-request output TPS ($\overline{\mathrm{TPS}}_\mathrm{req}$) against per-GPU system throughput ($\mathrm{TPS}_\mathrm{sys} / N_\mathrm{GPU}$), comparing BF16, INT4, and INT4+BDR (R128, key-only).

\begin{figure}[t]
    \centering
  \includegraphics[width=0.5\linewidth]{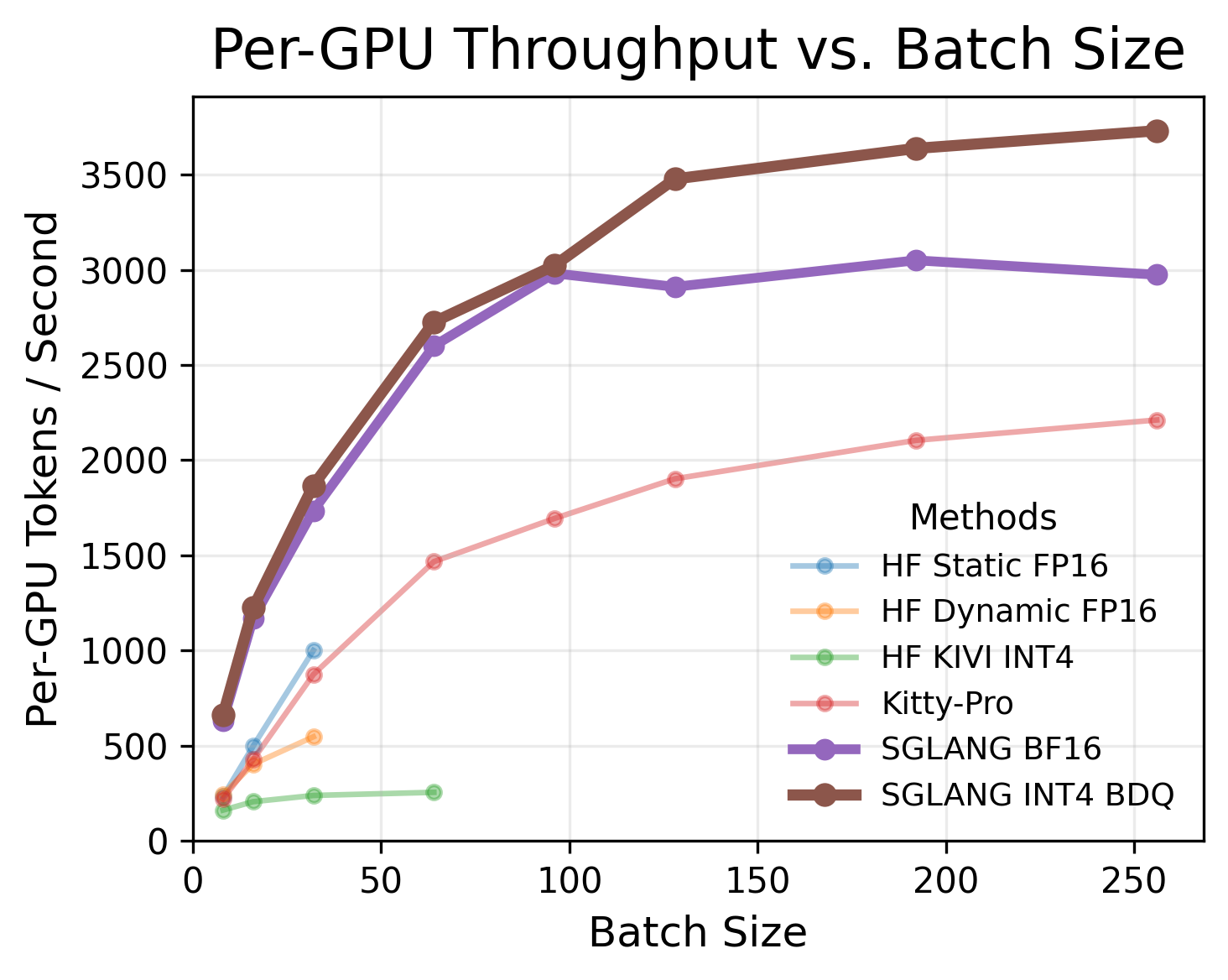}
  \caption{Qwen3-8B per-GPU throughput vs.\ batch size. More complex methods (e.g., Kitty) do not match the performance of BF16 or INT4+BDR. INT4+BDR performs substantially better at larger batch sizes. Input $\approx$100 tokens, output $\approx$8000 tokens.}
  \label{fig:comparison-all}
\end{figure}